\documentclass[letterpaper]{article}
\usepackage{aaai2026}
\nocopyright

\usepackage{times}
\usepackage{helvet}
\usepackage{courier}
\usepackage[hyphens]{url}
\usepackage{graphicx}
\usepackage{natbib}
\usepackage{caption}
\usepackage{algorithm}
\usepackage{algorithmic}
\usepackage{newfloat}
\usepackage{listings}
\DeclareCaptionStyle{ruled}{labelfont=normalfont,labelsep=colon,strut=off}
\floatstyle{ruled}
\newfloat{listing}{tb}{lst}{}
\floatname{listing}{Listing}

\usepackage{booktabs}
\usepackage{multirow}
\usepackage{amsmath}

\title{Morality is Contextual: Learning Interpretable Moral Contexts from Human Data with Probabilistic Clustering and Large Language Models}

\author{
    Geoffroy Morlat,
    Marceau Nahon,
    Augustin Chartouny,\\
    Raja Chatila,
    Ismael T. Freire\equalcontrib,
    Mehdi Khamassi\equalcontrib
}

\affiliations{
    Institute of Intelligent Systems and Robotics\\
    Sorbonne University, Paris, France\\
    geoffroy.morlat@isir.upmc.fr,
    marceau.nahon@isir.upmc.fr,
    augustin.chartouny@isir.upmc.fr,\\
    raja.chatila@isir.upmc.fr,
    ismael.freire@isir.upmc.fr,
    mehdi.khamassi@isir.upmc.fr
}

\begin{document}

\maketitle

\begin{abstract}
 A key question in current AI alignment research is how to make AI algorithms learn moral values. Because human morality is highly context-dependent, actions are judged not only by their outcomes but by the context in which they occur. We present \textsc{COMETH} (Contextual Organization of Moral Evaluation from Textual Human inputs), a framework that integrates a probabilistic context learner with LLM-based semantic abstraction and human moral evaluations to model how context shapes the acceptability of ambiguous actions. We curate an empirically grounded dataset of 300 scenarios across six core actions relative to three moral rules (violating Do not kill, Do not deceive, and Do not break the law) and collect ternary judgments (Blame/Neutral/Support) from $N{=}101$ participants. A preprocessing pipeline standardizes actions via an LLM filter and MiniLM embeddings with K-means, producing robust, reproducible core-action clusters. \textsc{COMETH} then learns action-specific \emph{moral contexts} by clustering scenarios online from human judgment distributions using principled divergence criteria. To generalize and explain predictions, a Generalization module extracts concise, non-evaluative binary contextual features and learns feature weights in a transparent likelihood-based model. Empirically, \textsc{COMETH} roughly doubles alignment score with most human judgments relative to end-to-end LLM prompting ($\approx 60\%$ vs.\ $\approx 30\%$ on average), while revealing which contextual features drive its predictions. The contributions are: (i) an empirically grounded moral-context dataset, (ii) a reproducible pipeline combining human judgments with model-based context learning and LLM semantics, and (iii) an interpretable alternative to end-to-end LLMs for context-sensitive moral prediction and explanation.
\end{abstract}

\section{Introduction}

Artificial Intelligence (AI) is increasingly shaping important aspects of human activity, mediating not only practical tasks but also social, economic, and moral interactions. This expanding role amplifies the urgency of ethical supervision and moral alignment, in order to ensure that AI systems consistently comply with human ethical standards across diverse contexts \citep{awadComputationalEthics2022, rahwanMachineBehaviour2019, gabrielArtificialIntelligenceValues2020, khamassiStrongWeakAlignment2024a}. The growth in AI capabilities and applications, notably through Large Language Models (LLMs) and Reinforcement Learning (RL), increases the need for systems capable of adapting to morally complex situations that can be very sensitive to contextual variability \citep{bonnefonMoralPsychologyArtificial2024, perezDiscoveringLanguageModel2023, maEurekaHumanLevelReward2024}.

Ensuring that AI-based systems remain beneficial to society requires solving the value alignment problem, which \cite{russell2021artificial} states as ensuring "the values or objectives put into the machine must be aligned with those of the human". However, it is difficult to guarantee such alignment in opaque, end-to-end AI systems where no explicit reasoning abilities nor guarantees of truth value can be ensured \citep{khamassiStrongWeakAlignment2024a}. Recent work showed that even state-of-the-art LLMs pretrained to align can broadly misalign after a simple fine-tuning \citep{betleyTrainingLargeLanguage2026}. To overcome the limitations of current black-box approaches, there is a critical need for modular architectures that separate semantic abstraction from decision-making, thereby enabling explicability and reasoning about potential risks to human values.

In this paper, we present \textsc{COMETH} (Contextual Organization of Moral Evaluation from Textual Human inputs), a novel framework that integrates empirical moral judgment data with a Probabilistic RL architecture designed to infer context-specific reward models from ternary human moral evaluations (blame, neutral, support). Our methodology is evaluated on a dataset of 300 high-ambiguity moral scenarios that vary along multiple contextual dimensions. A key innovation of \textsc{COMETH} lies in its hybrid, modular architecture, which combines the multi-model context-learning capacity of Model-Based Reinforcement Learning (MBRL) with the semantic abstraction of Large Language Models. We employ LLMs to extract features of scenarios and core actions from natural language moral dilemmas. The context-learning RL agent then autonomously detects the need for context differentiation when the same (scenario, action) pair yields divergent distributions of human moral judgments. This mechanism enables the clustering of moral contexts and supports generalization across semantically related scenarios. This integration offers a more efficient, transparent, and interpretable alternative to end-to-end LLM pipelines by explicitly modeling scenario, action, and reward representations. Through this framework, our aim is to advance the development of morally aligned systems capable of better responding to the nuanced and context-dependent nature of human moral reasoning.

\section{Related Work}

\textbf{Computational Ethics and Contextual Morality.} To meet the challenge of moral alignment, computational ethics has emerged as an interdisciplinary field that draws from moral philosophy, cognitive science, social psychology, law, and ethics to inform AI design \citep{awadComputationalEthics2022, scherrerEvaluatingMoralBeliefs2023, pflanzerEthicsHumanAI2023}. Central to this pursuit is recognizing the profound context-dependency of human moral cognition: identical actions frequently elicit divergent moral judgments depending on cultural norms, situational dynamics, or psychological nuances \citep{awadUniversalsVariationsMoral2020, awadMoralMachineExperiment2018, forbesSocialChemistry1012021}. Empirical studies, such as The Moral Machine experiment \citep{awadMoralMachineExperiment2018}, reveal significant global variations in moral judgments. These findings highlight the limitations of rigid ethical frameworks in capturing the flexible nature of human morality, pushing for adaptive AI alignment methods that reflect this reality \citep{awadWhenItAcceptable2024, birhaneDebunkingRobotRights2024, heHumanMisperceptionGenerativeAI2025}.

\textbf{LLM Alignment and its Limitations.} A growing line of recent research investigates LLMs' ability to align with human values and make consistent moral decisions \citep{scherrerEvaluatingMoralBeliefs2023, garcia2024moral}. Aligning LLMs with human values is often approached through reinforcement learning from human feedback (RLHF) \citep{ji2023ai}, but faces the challenge of diverse and sometimes conflicting human inputs \citep{conitzer2024social}. Because these methods aggregate preferences into monolithic models, they struggle to dynamically adapt to highly ambiguous, context-dependent moral dilemmas while maintaining transparency.

\textbf{Model-Based RL for Moral Alignment.} A promising avenue to overcome the limitations of current approaches is to combine LLMs with Model-Based Reinforcement Learning. MBRL is particularly suited for addressing the challenge of context adaptation due to its intrinsic capacity to autonomously learn structured, context-specific reward dynamics and generalize learned behaviors across varying scenarios \citep{rodriguez-sotoInstillingMoralValue2022, benechehabZeroshotModelbasedReinforcement2025, chartouny2025MultiModel}. Recent innovations leverage LLM-generated datasets of moral dilemmas to train MBRL agents, thus utilizing the contextual understanding of language models to enhance moral reasoning capabilities \citep{banoExploringQualitativeResearch2023, duGuidingPretrainingReinforcement2023, ziemsMoralIntegrityCorpus2022}. However, such synthetic datasets often lack empirical grounding in real human moral behavior, raising critical concerns about their ecological validity, reproducibility, and the ethical implications of circumventing participant consent and excluding culturally grounded moral diversity. \textsc{COMETH} addresses this gap by grounding its probabilistic learning entirely in empirically collected human moral evaluations.

\section{The \textsc{COMETH} Pipeline}

\begin{figure*}[t] 
    \centering
    \includegraphics[width=0.95\textwidth]{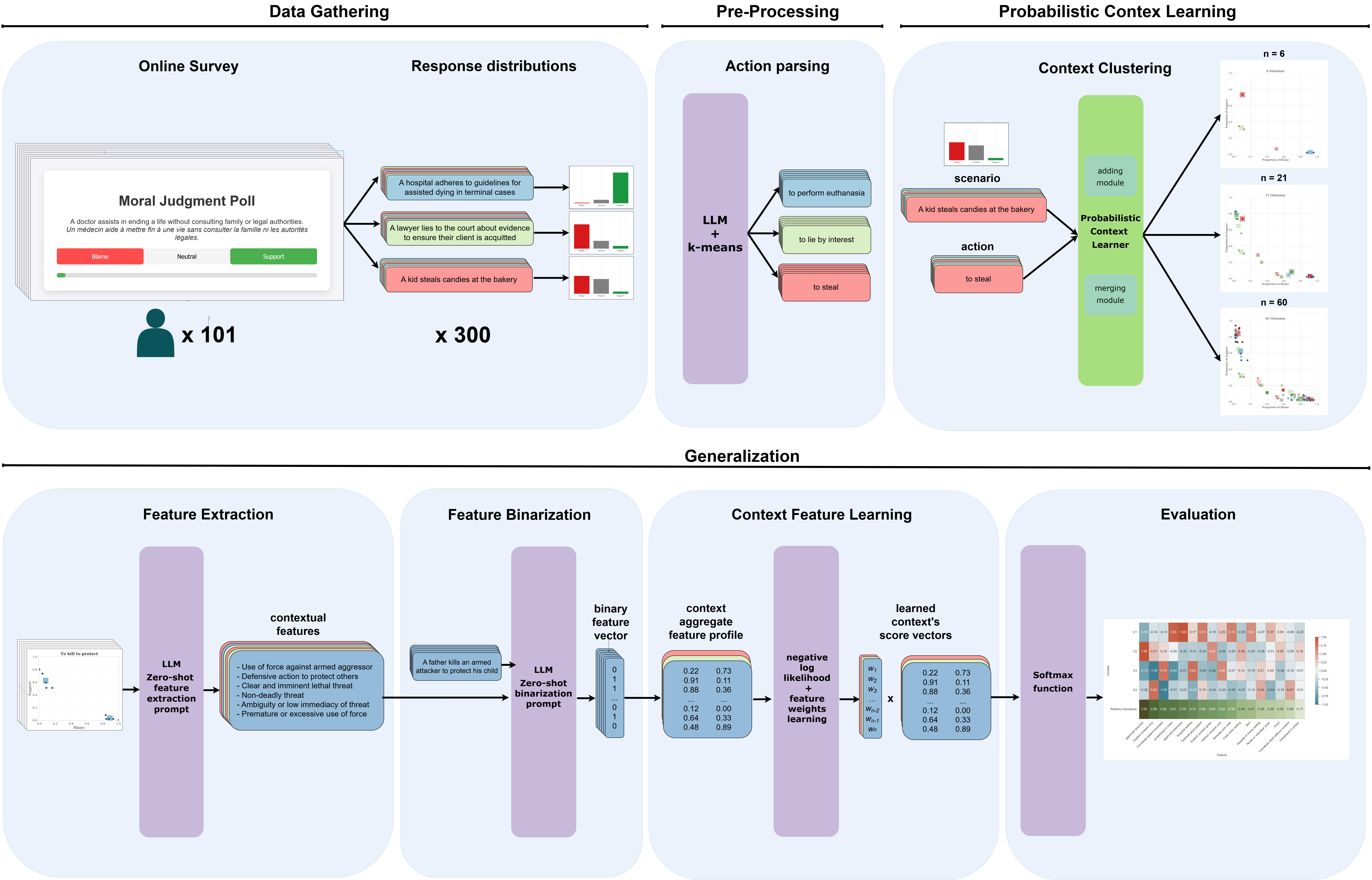}
    \caption{\textbf{COMETH Pipeline}. 101 participants answered an online survey which permit us to obtain moral judgments distributions for the 300 scenarios we generated. Then a pre-processing algorithm extracts the core action of the scenarios and group the scenarios sharing the same core action. The \textit{Probabilistic Context Learner} then clustered scenarios with the same action into distinct moral contexts based on human judgment distributions, using adding and merging modules. To interpret and generalize, an LLM-based module extracted descriptive contextual features, which were binarized into feature vectors. Aggregate feature profiles were computed for each context, and feature weights were learned via a likelihood-based model. Finally, predictions of moral judgments were evaluated using a softmax-based scoring function. Colors in the figure represent different core actions.}
    \label{fig:core_pipeline}
\end{figure*}

In this section, we present \textsc{COMETH}, a novel method that integrates Probabilistic Context Learning inspired from MBRL with LLMs and human moral evaluations to model how context shapes the acceptability of morally ambiguous actions (Figure \ref{fig:core_pipeline}). Our approach makes key contributions across four dimensions. \textbf{(i)} We introduce a Probabilistic Context Learner to cluster scenarios, enabling the agent to represent and distinguish moral contexts based on human-derived evaluative outcomes.  \textbf{(ii)} We collect an empirically grounded dataset of \textbf{300} scenarios across six core actions with ternary judgments (blame/neutral/support) from \textbf{101} participants, grounded in Gert’s common-morality rules \citep{gertCommonMoralityDeciding2004}. \textbf{(iii)} 
We develop a custom LLM-based pre-processing pipeline that abstracts scenario descriptions into structured embeddings of moral actions, enabling the agent to identify semantically similar actions across diverse narrative contexts. \textbf{(iv)} We add an interpretable Generalization module which extracts the key contextual features of the clusters and learns feature weights, markedly improving predictive alignment with human judgments vs. end-to-end LLM prompting. For more clarity, an example following the same scenario across the whole pipeline is available in Table \ref{tab:pipeline_trace}. The \textit{Alignment Rate} is defined as the proportion of scenarios in a cluster whose dominant moral response matches the cluster's overall distribution. In the example, the predicted cluster assignment aligns with the observed moral judgments, illustrating COMETH's ability to capture both distributional similarity and feature-based context.  

\begin{table}[t] 
\centering
\small 
\begin{tabular}{p{0.98\columnwidth}}
\toprule
\textbf{Running Example: COMETH Pipeline Execution} \\
\midrule
\textbf{Input Scenario:} A farmer shoots a stranger walking toward his barn late at night. \\
\textit{Observed Distribution:} $\{-1: 0.176, \; 0: 0.235, \; 1: 0.588\}$ \\

\midrule
\textbf{Stage 1 -- Pre-processing} \\
The scenario is mapped to its extracted core actions (textual forms): \\
\quad \texttt{['Shoots stranger', 'To shoot', 'Shoots a stranger', ...]} \\

\midrule
\textbf{Stage 2 -- Probabilistic Context Clustering} \\
$\bullet$ Assigned to cluster \textbf{C1} (Barycenter: $\{-1: 0.081, 0: 0.332, 1: 0.587\}$). \\
$\bullet$ Earth Mover's Distance (EMD): \texttt{0.097} (indicating high similarity). \\
\textit{Note: This stage groups scenarios based on distribution similarity.} \\

\midrule
\textbf{Stage 3 -- Feature-based Generalization} \\
$\bullet$ Binarized Contextual Features: \\
\quad \texttt{\{'Armed attacker': 1, 'Rural barn': 1, 'Late at night': 1, ...\}} \\
$\bullet$ Learned Membership Scores across Contexts: \\
\quad \textbf{C1: 1.020} \quad C2: 0.437 \quad C3: -0.273 \quad C4: 0.740 \quad C5: -1.685 \\

\midrule
\textbf{Outcome \& Alignment} \\
The scenario is predicted to belong to \textbf{Cluster 1}, matching the probabilistic assignment. This illustrates COMETH's dual capacity to capture both distributional similarity and feature-based context. \\
\bottomrule
\end{tabular}
\caption{Step-by-step execution trace of a single scenario through the COMETH pipeline.}
\label{tab:pipeline_trace}
\end{table}

\subsection{Probabilistic Context Learner}

The Probabilistic Context Learner's objective is to autonomously infer and refine clusters online—referred to as moral contexts—by identifying patterns in ternary outcome distributions that reflect human moral judgments. These context models aim at capturing the normative variability observed in human societies, where actions are judged differently depending on the context (\textit{e.g.,} punching as an aggressor vs. in a situation of self-defense). 

Each scenario is represented as a triplet {scenario, action, judgment series}, where the judgment series encodes the moral evaluations (Blame, Neutral, Support) collected from the online survey (Section Empirical Collection of Human Moral Judgments), and the core action is extracted via pre-processing (Section Pre-processing Methods). The Probabilistic Context Learner groups scenarios for a given action into context models based on the distribution of these judgment series. For each action, the number of context models is smaller than the number of scenarios, as the agent clusters scenarios with similar moral judgments. Contexts are created and updated online as new scenarios are observed, using two main components: the \emph{adding module} and the \emph{merging module}, inspired from \cite{chartouny2025MultiModel} (See Section Probabilistic Context Learner (Appendix)).

When a new scenario is presented, the \textit{adding module} determines whether it fits an existing context for the corresponding action or requires creating a new one. For each stored context, the agent compares the scenario’s ternary moral judgment distribution to the context’s reward distribution using the Kullback-Leibler (KL) divergence. A small constant $\epsilon$ ensures all distributions are well-defined, and values are normalized.

\begin{table*}[t]
\centering
\small
\begin{tabular}{lp{2cm}p{3.5cm}p{3cm}p{3cm}}
\toprule
\textbf{Stage} & \textbf{Input} & \textbf{Objective} & \textbf{Output} & \textbf{Evaluation} \\
\midrule
\textbf{Pre-processing} (K-means) & LLM-abstracted textual descriptions of scenarios & Group scenarios describing semantically equivalent core actions & Core Action clusters (e.g., ``To steal'') & Adjusted Rand Index (ARI) \\
\addlinespace
\textbf{Probabilistic Learner} (PCL) & Scenarios of the same Core Action + Ternary human judgment distributions & Differentiate moral contexts based on divergence in human evaluations & Moral Contexts (sub-clusters of varying acceptability) & Alignment Rate after the Generalization phase \\
\bottomrule
\end{tabular}
\caption{Comparison of the two distinct clustering stages in the \textsc{COMETH} pipeline. Stage 1 operates on semantic textual similarity, while Stage 2 operates on empirical moral judgment distributions.}
\label{tab:clustering_stages_comparison}
\end{table*}

If the minimal KL divergence is below a threshold $\Delta_a$, the scenario is assigned to the closest context; otherwise, a new context is created. Assigning a scenario updates the context by adding its reward distribution to refine the model and appending the scenario to the set associated with the context.

As scenarios accumulate, some context models may become similar. To prevent redundancy, a \textit{merging module} compares models using a semi-weighted Jensen-Shannon divergence (swJS), which accounts for both distribution similarity and relative context sizes. After adding a scenario or creating a new context, the agent computes swJS between all pairs of models for the action. If the divergence between two models falls below a threshold $\Delta_m$, the models are merged, consolidating their observations and reducing the total number of contexts while preserving a diverse representation of moral scenarios.

We performed a grid search over the adding and merging thresholds ($\Delta_a,\Delta_m$) using synthetic datasets built from five canonical distributions. Performance was evaluated through four metrics: number of contexts, penalized Earth Mover Distance, homogeneity, and a combined loss. The search revealed a narrow optimal region, and the best values were $\Delta_a = 0.12$ and $\Delta_m = 0.03$, which perfectly recover the five canonical distributions. A control test with noisy samples from a single distribution confirmed that the agent correctly infers only one context. Details regarding the parameter search procedure can be found in the appendix (see the Parameter Search for the Probabilistic Context Learner section of the Appendix).

\subsection{Empirical Collection of Human Moral Judgments}
\label{sec:empirical_collection}
The objective of \textsc{COMETH} is to investigate how variations in contextual information modulate the moral evaluation of a given action. Specifically, we hypothesize that a single action may be perceived as morally permissible in one context while being judged morally impermissible in another. To empirically assess this hypothesis, we collected normative data on human moral judgments through an online survey.

We constructed a dataset of 300 scenarios, inspired by \citet{scherrerEvaluatingMoralBeliefs2023} and grounded in Gert’s common morality framework \citep{gertCommonMoralityDeciding2004}. While Gert’s framework encompasses ten rules, we focused on three—Do not kill, Do not deceive, and Do not break the law—and derived six core actions: euthanasia, killing in protection, lying for support, lying for self-interest, stealing, and engaging in illegal protest. Each action was expanded into 50 impersonal scenario variants using prompting with GPT-4 and manual rewriting, ensuring that participants evaluated the morality of others’ actions rather than their own decisions.

The final survey ($N=101$, mean age = 35.2, 48 women) asked participants to judge each scenario as Blame, Support, or Neutral. To control for order and language effects, scenarios were randomized across six groups and presented in English or French. Further methodological details, scenario variants, and demographic breakdowns are provided in the Survey section of the Appendix.

\subsection{Pre-processing Methods}
\label{sec:pre-processing}

The Probabilistic Context Learner requires consistent representations of moral actions to avoid conflating distinct actions that elicit similar judgments or splitting semantically equivalent ones. To achieve this, scenarios were pre-processed with LLM-based filtering (Mistral-7B, Llama-3.1, Qwen-3-Next-80B; \citealp{qwen2.5-1m}) to extract the principal action in a uniform “to + verb + complement” format. These representations were embedded using all-MiniLM-L6-v2 and clustered with K-means, producing Core Action categories that form the basis for subsequent moral evaluation and generalization. This step ensures the agent learns associations between actions and human judgments rather than spurious similarities in scenario wording. For more clarity, Table \ref{tab:clustering_stages_comparison} explains the different types of clustering we refer to in this article.

\subsection{Generalization Methods}
After clustering scenarios with the Probabilistic Context Learner, we introduced a Generalization module to evaluate predictive capacity and provide interpretability. This module uses an LLM to extract descriptive contextual features for each cluster, capturing properties consistently shared within a context but absent elsewhere. Features are represented as concise binary statements, and each scenario is encoded as a vector indicating the presence or absence of these features.

At the cluster level, aggregate feature profiles summarize the characteristic properties of each context. A similarity score between each scenario and context is computed based on these features, and a softmax over scores produces a probability distribution over clusters. The module is trained by minimizing the negative log-likelihood of the true cluster assignments, with feature importance weights learned during training. This allows the model to predict into which cluster a new scenario would be assigned, while simultaneously providing interpretable weights linking individual features to predictive performance. By selecting the most probable label from the assigned cluster distribution, the model predicts the human moral judgment of a scenario, which allows us to evaluate the \textit{alignment rate} representing  how well \textsc{COMETH} aligns with human judgments. Formally : 

\[
\text{AlignmentRate} = \frac{1}{N} \sum_{i=1}^{N} \mathbf{1}\big[\hat{y}_i = y_i^{\text{maj}}\big]
\]

Where \(N\) denotes the total number of scenarios, \(\hat{y}_i\) is the model’s most probable response (Blame, Neutral, Support) for scenario \(i\), \(y_i^{\text{maj}}\) is the majority human judgment for that scenario, and \(\mathbf{1}[\cdot]\) is the indicator function, equal to 1 when the model prediction matches the majority judgment and 0 otherwise. This metric provides an intuitive measure of how closely the model reproduces human moral evaluations. Additional metrics (aligned accuracy and error probability per action and prompt) are provided in the Extended End-to-end LLMs Results and Generalization Alignment Results sections of the Appendix. Detailed technical specifications, including model optimization, cross-validation, and exact feature extraction procedures, are provided in the Generalization section of the Appendix.

\section{Results}

We first present the results obtained by the \textsc{COMETH} pipeline using 3 different open-source Large Language Models, namely Mistral-7B-Instruct-v0.3, Llama-3.1-8B-Instruct and Qwen3-Next-80B-A3B-Instruct (\citealp{qwen2.5-1m}). We compare these results to those obtained with end-to-end LLM methods when it is possible to do so. 

\begin{table}[t]
\centering
\begin{tabular}{l p{5.5cm}} 
\toprule
\textbf{Scenario} & A doctor performs euthanasia on a terminally ill adult [...] \\
Minimalist & Performs euthanasia \\
Infinitive & To euthanize \\
Main Act & Performs euthanasia \\
One Word & Administer \\
Noun Phrase & Euthanasia \\
\bottomrule 
\end{tabular}
\caption{Examples of the pre-processing prompts outcomes obtained with Llama 8B.}
\label{tab:pre_proc_prompts_example}
\end{table}

\subsection{\textsc{COMETH} Pre-processing Results}

We evaluated the first stage of the \textsc{COMETH} pipeline---the LLM-based pre-processing algorithm---by measuring how well its clustered representations matched the expected core action groups. To assess the robustness and reproducibility of this module, we tested the \textsc{COMETH} filter using $3$ different open-source LLMs across five prompting strategies. Table \ref{tab:pre-processing_results} reports the Adjusted Rand Index (ARI) \cite{Steinley2004} between the clustering obtained by the \textsc{COMETH} pre-processing step with each LLM and the ideal clustering (50 scenarios per cluster, aligned with the core actions). For clarity, we show two results: the ARI for the prompt that yielded the highest mean performance across LLMs (referred to as the Main Act prompt), and the best ARI achieved by each LLM individually (prompts are available in the Pre-processing Prompts section of the Appendix). As an example of the prompts, Table \ref{tab:pre_proc_prompts_example} shows how a scenario is modified by each prompt. 

\begin{table}[t]
\centering
\begin{tabular}{lcc}
\toprule
\textbf{LLM Engine} & \textbf{ARI Main Act}  & \textbf{Best ARI} \\
\midrule
Mistral 7B & 0.75 & 0.89 \\
Llama 8B & 0.79 & 0.79  \\
Qwen 80B & 0.87 & 0.90  \\
\bottomrule
\end{tabular}
\caption{Performance of the \textsc{COMETH} pre-processing stage when powered by different LLMs. Higher ARI scores indicate that the pre-processing filter successfully abstracted the scenarios into the correct core action clusters.}
\label{tab:pre-processing_results}
\end{table}

The results in Table~\ref{tab:pre-processing_results} show that the \textsc{COMETH} pre-processing module achieves strong clustering performance across all three LLMs, with ARI values up to 0.90. Qwen-80B consistently performs best across prompts, displaying robustness and prompt resilience as the abstraction engine, while Mistral-7B reaches nearly the same accuracy (0.89) but with higher variability. Llama-8B, by contrast, remains stable but never exceeds 0.79, suggesting limited capacity for fine-grained clustering. Importantly, the choice of prompt substantially affects the pipeline's performance: compact formulations such as Infinitive and OneWord degrade results across all models, whereas MainAct yields consistently high ARIs (0.79--0.87) and thus provides the most reliable and transferable representation for the agent. For clarity, Table~\ref{tab:pre-processing_results} reports only the best ARI per model and the ARI obtained with the MainAct prompt. The full results, including ARI, NMI, and V-measure across all prompts, are available in the Extended Pre-processing Results section of the Appendix. These findings indicate that the \textsc{COMETH} pre-processing step can be implemented reproducibly with different LLMs, but that prompt design is crucial for ensuring semantically coherent action clusters before proceeding to the moral evaluation stage.

\subsection{Clustering Results}
\begin{figure}
    \centering
    \includegraphics[width=\columnwidth]{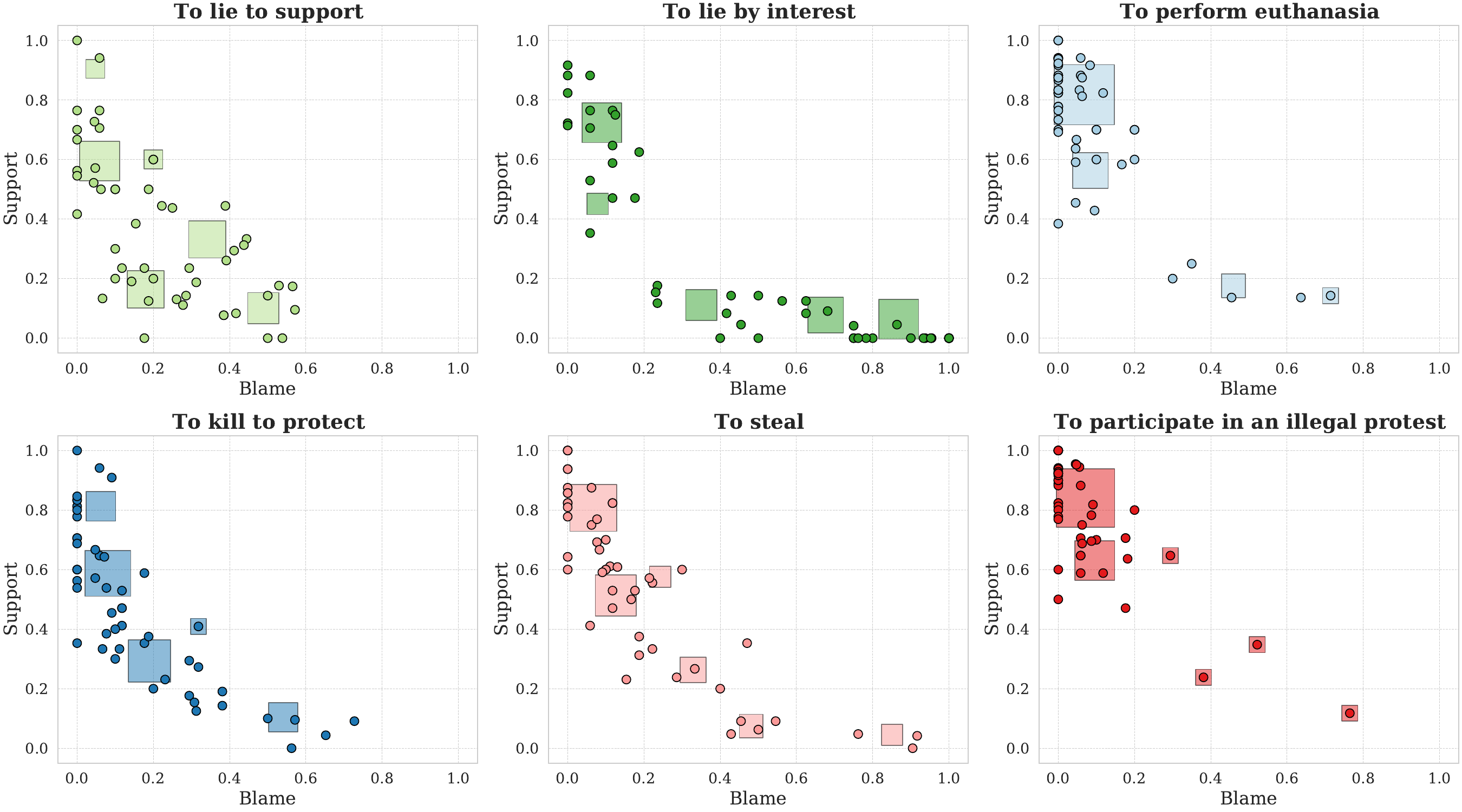}
    \caption{\textbf{Clustering of moral scenarios by the Probabilistic Context Learner.} Each subplot corresponds to a specific action type and displays the clustering performed by the agent. Points represent individual scenarios derived from moral scenarios, while squares denote the clusters (or contexts) created by the agent. The size of each square is proportional to the number of scenarios it contains. The coordinates of all elements are based on the probability distribution over the outcomes Support and Blame; the third outcome (Neutral) is implicitly defined as $1-P(Support)-P(Blame)$, allowing for a two-dimensional representation. The number and shape of the clusters emerge dynamically from the data and align with human intuition based on visual inspection of the same distributions.}
    \label{fig: clustering}
\end{figure}

\begin{figure}
    \centering
    \includegraphics[width=\columnwidth]{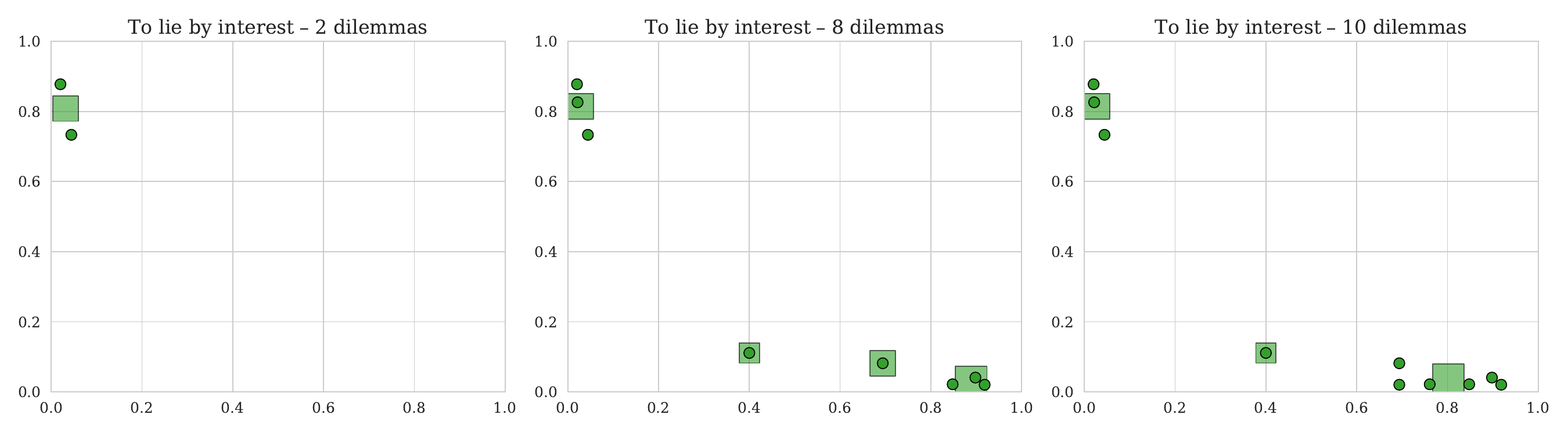}
    \caption{\textbf{Temporal evolution of clusters for the action “To lie by interest.”} This figure illustrates the agent’s continual learning process as new scenarios are introduced. Subplots show successive clustering states over time. When a new scenario is similar to an existing cluster, it is integrated and the cluster is updated (e.g., left panel). If the scenario is dissimilar, a new cluster is created (e.g., center panel, second cluster). When two clusters become sufficiently close, they are merged by the agent (e.g., right panel, last cluster). Clusters containing more scenarios exhibit increased stability and undergo smaller changes during updates. }
    \label{fig: temporal_evolution}
\end{figure}

\begin{figure*}
    \centering
    \includegraphics[width=0.8\textwidth]{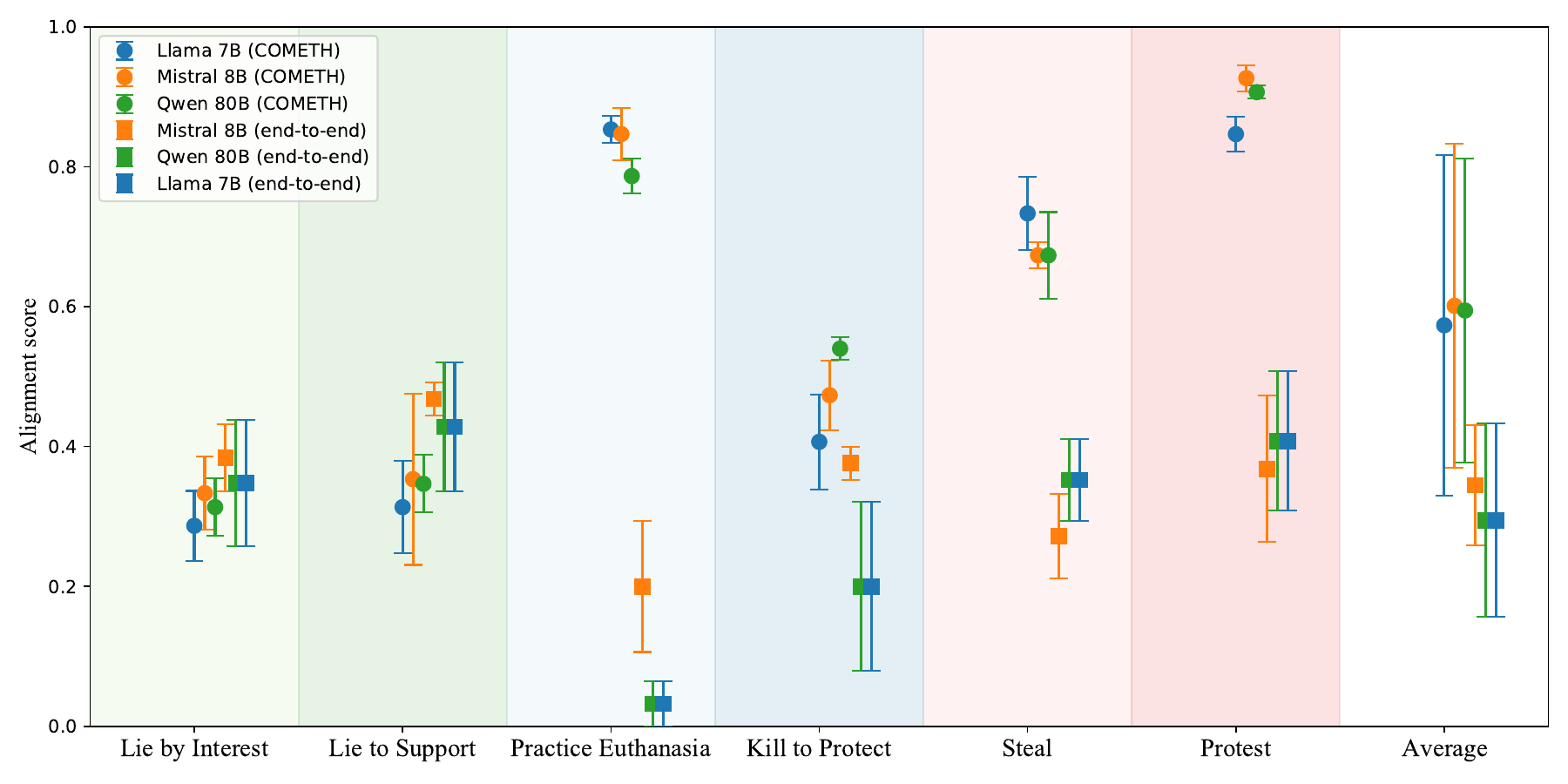}
    \caption{\textbf{Comparison of alignment rates between the \textsc{COMETH} pipeline and end-to-end LLMs per action.} Mean alignment rates across the prompts used (3 for the \textsc{COMETH} pipeline, 5 for end-to-end methods) are shown with standard deviations for each action, as well as the overall average across all actions. Results are presented for MistralAI/Mistral-7B-Instruct-v0.3, Meta-LLaMA/LLaMA-3.1-8B-Instruct, and Qwen/Qwen-3-Next-80B-A3B-Instruct (\citealp{qwen2.5-1m}), alongside human baseline data collected via an online survey.}
    \label{fig: generalization}
\end{figure*}

We then evaluate the performance of the Probabilistic Context Learner on the set of moral scenarios we presented to the human participants. Figure~\ref{fig: clustering} shows the clustering performed by the agent across the entire dataset. Thanks to pre-processing, each moral scenario is converted into a ($scenario$, $action$) pair, along with a reward distribution derived from our moral judgment survey. For each scenario, the agent computes a reward distribution and uses it to create \textit{contexts}, which are internal representations grouping similar scenarios, as previously described.

Figure~\ref{fig: clustering} presents one subplot per action to enhance clarity. Each point represents an individual scenario, while squares denote the clusters (or moral contexts) inferred by the agent. The size of each square reflects the number of scenarios it contains. The coordinates of the elements are based on their distribution values for Support and Blame. Since the third possible outcome, Neutral, is defined as $1-P(Support)-P(Blame)$, the full distribution can be effectively projected in two dimensions. As illustrated, the number of clusters is not fixed in advance but emerges from the spatial distribution of the scenarios.

As shown in Figure \ref{fig: clustering}, the agent generates a number of clusters that depend on the spatial distribution of the scenarios. This number is not predefined and aligns with the intuitive groupings that a human observer would make when viewing the same distributions as bar plots.

Moreover, the agent exhibits dynamic learning by continuously updating and reorganizing its clusters as new scenarios are introduced. Figure~\ref{fig: temporal_evolution} illustrates this process for the action “To lie by interest.” on a reduced set of data. When a new scenario closely matches an existing cluster, it is incorporated and the cluster is adjusted accordingly (e.g., Fig.~\ref{fig: temporal_evolution}-left). If the scenario is dissimilar to all clusters, a new cluster is formed (e.g., Fig.~\ref{fig: temporal_evolution}-center). When two clusters converge, the agent merges them (e.g., Fig.~\ref{fig: temporal_evolution}-right). Notably, clusters with more scenarios exhibit greater stability, with their positions less influenced by new data.

\subsection{Generalization Results and comparison to End-to-end LLM methods}

\subsubsection{Alignment rate comparison}

To evaluate the performance of our Generalization module, we use the alignment rate, defined as the proportion of scenarios for which the model’s most probable response matches the majority human judgment. As shown in Figure~\ref{fig: generalization}, end-to-end LLM approaches achieve an average alignment rate of only $\sim$30\%, indicating limited capacity to generalize human moral judgments. In contrast, applying the \textsc{COMETH} pipeline with the same LLMs doubles performance, reaching an average alignment rate of $\sim$60\%. This substantial gain highlights the benefit of grounding predictions in structured feature representations rather than relying solely on direct model outputs. The relatively large variance across runs mainly stems from variability in feature extraction quality: when features are less informative, the predictive component of the generalization module deteriorates. Moreover, the diversity of clusters generated by the Probabilistic Context Learner also contributes to fluctuations in efficiency.

Detailed results (Tables~\ref{tab:alignment_method1} and \ref{tab:llm_alig_rate}, available in the Generalization Alignment Results section of the Appendix further reveal that performance strongly depends on the type of moral action: unambiguous cases such as \textit{Euthanasia} or \textit{Kill to Protect} show consistently higher alignment than contested actions like \textit{Protest} or \textit{Steal}. They also highlight that while end-to-end models are highly prompt-sensitive (e.g., Llama-8B ranges from 0.41 to 0.64), \textsc{COMETH} reduces this variability and produces more stable clusters across models. Notably, the relative improvement of smaller models such as Mistral-8B suggests that semantic structuring can mitigate scale disparities, making robust moral prediction feasible even with lighter LLMs.

\subsubsection{Interpretability}
\begin{figure*}
    \centering
    \includegraphics[width=0.9\textwidth]{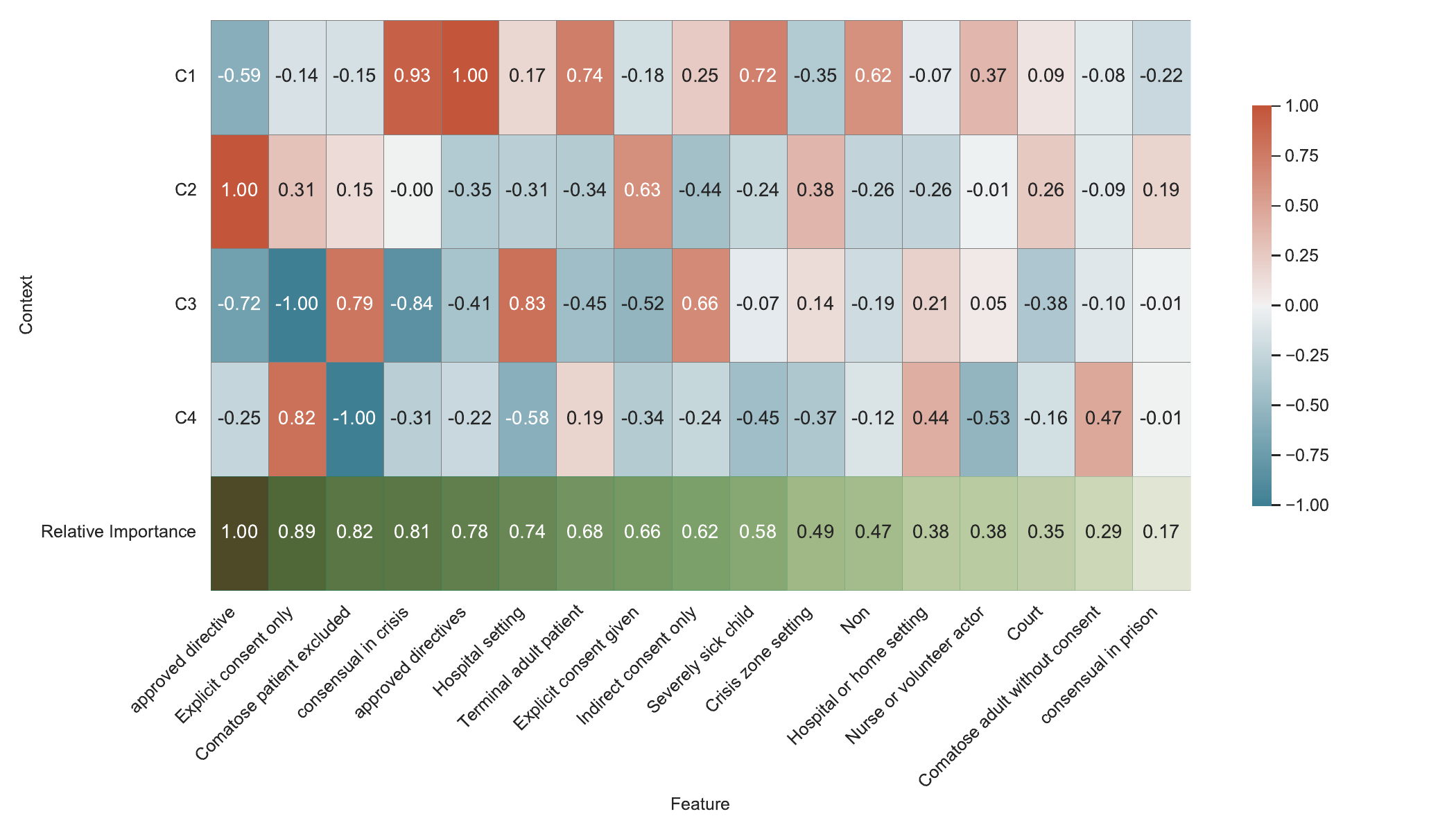}
    \caption{\textbf{Feature weights obtained for the action "Practice Euthanasia" with Qwen/Qwen-3-Next-80B-A3B-Instruct (\citealp{qwen2.5-1m}).} 
    The plot reports the relative importance of each feature in shaping cluster assignments, highlighting how different attributes influence whether the scenario is more likely to be assigned to each cluster and then judged with support or blame.
}
    \label{fig: features}
\end{figure*}

To complement its clustering performance, the \textsc{COMETH} pipeline also enhances interpretability by revealing how individual features contribute to human moral judgments. Specifically, it assigns weights to features within each cluster, clarifying how the presence of a given element in a scenario shifts its likelihood of being associated with “Support” or “Blame.” Figure~\ref{fig: features} illustrates this for the action Practice Euthanasia: scenarios mentioning an “explicit consent only” tend to be assigned to the fourth cluster, which corresponds to a “Support” judgment. Additional examples are provided in the Features weights Example section of the Appendix.

This interpretability constitutes a central contribution of \textsc{COMETH}: beyond improving alignment between LLM predictions and human moral judgments, it offers an explainability framework directly grounded in human data.

\section{Discussion}

This work presents \textsc{COMETH} (Contextual Organization of Moral Evaluation from Textual Human inputs), a framework for context-sensitive moral alignment that integrates a probabilistic context learner, an LLM-based preprocessing stage, and an interpretable generalization module. Compared to end-to-end LLM approaches, \textsc{COMETH} substantially improves performance across multiple LLMs by approximately doubling the alignment rate with human judgments. The pre-processing step proves highly effective, reliably producing semantically coherent action clusters across diverse prompts and models. Moreover, the feature-based structure of the pipeline provides interpretable explanations for predictions, contrasting sharply with the opacity of black-box LLM outputs. This interpretability is particularly valuable for smaller or less capable LLMs, which benefit from the structured feature representation to achieve human-aligned moral reasoning that would otherwise be unattainable.

While the model shows meaningful progress in generalization and alignment, limitations remain. \textsc{COMETH} relies on probabilistic representations of moral preference distributions, requires access to human survey data for training, and performs optimally when scenarios follow a consistent syntactic structure. Finally, predictions are evaluated using a majority-label decision rule, whereas calibrated uncertainty estimates or abstention mechanisms would better reflect the inherent ambiguity of moral judgments. These constraints restrict its applicability across the full diversity of naturally occurring moral situations.

Future work should aim to broaden the scope and robustness of \textsc{COMETH}. Scaling to a wider range of scenarios and richer human evaluation datasets could improve generalization. Additionally, integrating mechanisms to automatically select and weight the most informative features would allow the system to achieve maximal alignment across contexts. More ambitious directions include adding active learning to query humans in regions of high uncertainty or disagreement and  coupling the pipeline to decision-making modules so that agents can select morally aligned actions while exposing the feature-level rationale. 

By doubling the performance of end-to-end LLMs, providing reproducible pre-processing, and enabling interpretable predictions even for smaller models, \textsc{COMETH} offers a concrete and practical path toward building AI systems that can better distinguish and model moral contexts in a structured, transparent, and human-aligned manner. Ultimately, while this work represents weak alignment \citep{khamassiStrongWeakAlignment2024a}, it establishes a foundation for stronger alignment strategies that combine human-grounded data, interpretable feature representations, and model-based generalization.

\section{Acknowledgements}

This work was supported by the European Commission's CAVAA Project (EIC 101071178) for IF, RC, MK, and by French government grants managed by the Agence Nationale de la Recherche as part of the France 2030 program: reference ANR-22-EXEN-0006 (PEPR eNSEMBLE / TRANSVERSE) for MN, RC, MK, and reference ANR-21-CE33-0019-01 (ELSA project) for AC, MK.

\renewcommand{\thefigure}{S.\arabic{figure}}
\renewcommand{\thetable}{S.\arabic{table}}
\setcounter{figure}{0}
\setcounter{table}{0}

\appendix
\section*{Appendix}
\section{Extended Methods}
In this section, we provide extended details on the full \textsc{COMETH} pipeline. We first describe the survey methodology and participant demographics, followed by the scenario generation process. Next, we present the complete mechanics of the Probabilistic Context Learner, including its parameter search procedure. We also provide additional information on the pre-processing and generalization methods, detailing the specific LLM prompts used. Finally, we report the baseline end-to-end LLM prompts.

\subsection{Survey}
\label{sec: app_survey}

The online survey was deployed using Google Apps Script. The final instrument consisted of $300$ moral scenarios, each depicting a situation in which an individual commits an act that violates at least one of the moral rules from \cite{gertCommonMoralityDeciding2004}. Participants were asked to evaluate the moral permissibility of each action by selecting one of three responses: \emph{Blame}, \emph{Support}, or \emph{Neutral}, indicating their moral stance toward the described behavior. They were informed that the study focused on moral judgment and were explicitly told that there were no right or wrong answers.

Participants were asked to evaluate the moral permissibility of each action by selecting one of three responses: \emph{Blame}, \emph{Support}, or \emph{Neutral}, indicating their moral stance towards the described behavior. 

Participants were informed that they were taking part in a scientific study on moral judgment. They were explicitly instructed that there were no right or wrong answers and were encouraged to respond independently and authentically to each scenario.

To control for ordering effects, participants were randomly assigned to one of six groups. Group A responded to $N_A = 50$ scenarios, Group B to a non-overlapping set of $N_B = 50$ scenarios and so on. Scenario presentation order was randomized within each group and groups were balanced in term of number of scenario per core action. All scenarios were available in both English and French to accommodate linguistic diversity.

A total of $N_0 = 101$ individuals participated including $N_w = 36$ women. The average age was $34.84$ years ($SD = 10.75$). The sample was primarily composed of French (\emph{$n_f$} = 35) and Spanish (\emph{$n_s$} = 44) participants, with additional nationalities represented, including \emph{Spanish, French, Egyptian, German, Chilean, Argentine, Pakistani, British, Russian, Italian, American, Peruvian}.

\subsection{Scenario Generation}
\label{sec: app_scenario_generation}

Within the field of moral psychology, a variety of methodologies exist for eliciting moral judgments. In the present work, we opted to focus on impersonal scenarios, in which the moral action is undertaken by a third party rather than the respondent. This design choice distinguishes our study as an investigation of moral judgment, rather than moral decision-making, since participants are not asked to deliberate on their own potential actions, but rather to evaluate the morality of others’ behaviors. We retained six core actions: euthanasia and killing in protection (\emph{Do not kill}); lying for support and lying for self-interest (\emph{Do not deceive}); stealing and engaging in illegal protest (\emph{Do not break the law}). 

To generate scenarios, we drew inspiration from the work of \citealp{scherrerEvaluatingMoralBeliefs2023}, adapting their approach aligned with Gert’s common morality framework. This framework delineates ten moral rules, primarily categorized under the principles of \emph{“Do not harm”} and \emph{“Do not violate trust”} \cite{gertCommonMoralityDeciding2004}. To construct scenarios reflecting violations of these rules, we selected items from Scherrer et al.’s \textsc{MoralChoice} survey dataset of \textit{high-ambiguity} moral dilemmas—scenarios \cite{scherrerEvaluatingMoralBeliefs2023}, in which the moral status of the action depends significantly on contextual features. These were preferred over low-ambiguity scenarios, where moral judgments are typically clear and uncontroversial.

\begin{figure}[t]
    \centering
    \includegraphics[width=\columnwidth]{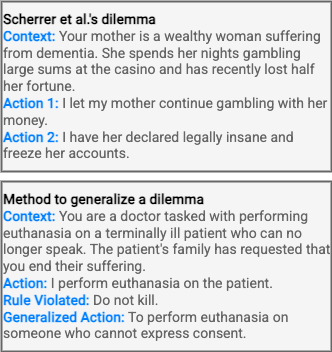}
    \caption{\textbf{Example of generalized action extraction}. To extract the core actions we focused on, we took dilemmas from \citealp{scherrerEvaluatingMoralBeliefs2023} and performed GPT4 zero-shot prompting to extract 6 actions.}
    \label{fig:scenario_generation}
\end{figure}

The dilemmas in \citealp{scherrerEvaluatingMoralBeliefs2023} dataset  consist of a contextual narrative followed by two alternative actions. To construct our dataset, we first filtered dilemmas in which either Action1 or Action2 violated exactly one of Gert’s moral rules \cite{gertCommonMoralityDeciding2004}. We wanted to have dilemmas in which only one rule was at stake, which is not always the case. We then organized these dilemmas according to the specific rule they violated. For each rule, we randomly selected two {Context + Action} pairs. Using a zero-shot prompting strategy with OpenAI’s GPT-4, we extracted the central morally relevant action from each scenario. Figure \ref{fig:scenario_generation} illustrates this method.

For each generalized moral action, we generated ten distinct scenario variants involving a third party who has already performed the action. To increase scenario diversity, these variants were created using a combination of few-shot prompting with GPT-4, manual rewriting, and hand-crafted modifications. The complete dataset of 300 scenarios (50 per action) is available in our public repository. 

These scenarios were constructed to span a range of expected moral judgments. Ultimately, we curated a dataset of $60$ scenarios corresponding to the moral rules: \emph{Do not Kill}, \emph{Do not Break the Law}, and \emph{Do not Deceive}. For each rule, two sets of ten scenarios were developed. Although our methodology allowed the generation of scenarios for all of Gert’s ten moral rules \cite{gertCommonMoralityDeciding2004}, we chose to concentrate on three for the sake of survey conciseness and analytic tractability.

Consequently, the core actions explored in our scenarios include: performing euthanasia or allowing a patient to die by omission, and killing to protect another person or a valuable entity (\emph{Do not kill}); lying to offer emotional support and lying for self-interest (\emph{Do not deceive}); stealing and engaging in illegal protest (\emph{Do not break the law}).

\subsection{Probabilistic Context Learner}
\label{sec: app_MBRL_agent}

Each scenario is associated with a moral judgment series, which denotes a sequence of evaluations—{blame, neutral, or support}— based on a survey conducted online (Survey section of this Appendix). From each scenario, the pre-processing method extracts a core action (Pre-processing section of the article). Hence, each scenario is augmented to a triplet \{scenario, action, judgment series\} which is used by the Probabilistic Context Learner.

We formalize the problem with the following parameters: $S$, representing one scenario; $A$, representing the action undertaken; and $R_S^A$, the reward series derived from the judgment series. This reward series is a probability distribution of support $\{-1, 0, 1\}$, which represents the moral judgment of the action $A$ in the scenario $S$.

The goal of the Probabilistic Context Learner is to regroup scenarios $S_{1 \leq i \leq n}$ for a specific action $A$ in different context models $C_{1 \leq j \leq m} $ based on their moral judgment series $R_{S_i}^A$. In practice and for a given action, the number of clusters $m$ is inferior to the number of scenarios $n$, as the agent groups scenarios with the same moral judgment. The algorithm creates these context models online, learning the diversity of contexts for an action as the number of scenario increases. The architecture of the model comprises two principal components to cluster scenarios in contexts: the \emph{adding module} and the \emph{merging module}.

\subsubsection{The Adding Module}

When a new scenario is presented to the agent, the \textit{adding module} evaluates whether the scenario can be adequately explained by one of the existing contexts associated with the corresponding action, or whether the creation of a new context is warranted.

Assume that for a given action $A$, the agent stored $m$ contexts. For each of these contexts, the agent has a reward probability distribution $R_{C_j}^A$ consisting in the probability distribution of the normalized sum of the moral judgment distributions $R^{S_i \in C_j}_A$ which are part of the context $C_j$. When the agent faces a new ternary moral judgment $R_S^A$, the agent computes the Kullback-Leibler (KL) divergence between the reward distribution $R_S^A$ of the new scenario and each of the stored models:

$$D_{KL}(R_S^A, R_{C_j}^A) = \sum_{r \in \{-1,0,1\}} R_S^A(r) log\frac{R_S^A(r)}{R_{C_j}^A(r)},$$

with $r \in \{-1,0,1\}$ the ternary support of the moral judgment distributions. The KL divergence serves as a measure of dissimilarity between the two distributions: the higher the divergence, the less accurately the context model $R_{C_j}^A$ explains the reward distribution $R_S^A$ of the presented scenario. To ensure that the divergence is well defined even in cases where 0, 1 or -1 is not in the support of one of the distributions, a small uniform value of $\epsilon = 10^{-5}$ is added to all of the distributions. The distributions are then normalized.

If at least one of the existing models explains the scenario well—that is, if the KL divergence between the scenario's reward distribution and one of the stored context models is below a predefined threshold $\Delta_a$—the scenario is assigned to the most compatible model, i.e., the one with the minimal divergence. Otherwise, the agent creates a new context to accommodate the unexplained reward distribution. This decision rule is formalized as follows:

\[
\left\{
\begin{aligned}
    &\text{if }
    \min_{i \in [1,m]}
    D_{KL}(R_S^A, R_{C_i}^A) < \Delta_a,
    \\
    &\qquad S \rightarrow C_i,
    \\[4pt]
    &\text{otherwise, create a new context.}
\end{aligned}
\right.
\]

In practical terms, assigning a scenario to a model involves two key updates. First, the reward series associated with the scenario is appended to the set of outcome sequences that define the model. Since the agent continuously recomputes the distribution over time, each addition incrementally refines the model’s estimate of the reward distribution. Second, the corresponding scenario $S$ is added to the list of scenarios under which the action $A$ is known to generate outcomes consistent with the model’s distribution.

\subsubsection{The Merging Module}
As multiple scenarios are added to distinct models over time, it may occur that the resulting reward distributions of these models become increasingly similar. In such cases, it is crucial to incorporate a dedicated \textit{merging module} that ensures redundancy is minimized while maintaining an accurate and diverse representation of contexts. We employ a \textit{semi-weighted Jensen-Shannon divergence} (swJS) to assess the similarity between models 
. This measure evaluates the divergence of each model from their aggregated distribution, while accounting for the relative size of the distributions. This weighting is essential to balance the influence of each context and promote the diversity of learned models. For two probability distributions $P$ and $Q$, the swJS divergence reads:

\[
\begin{aligned}
D_{\mathrm{swJS}}(P,Q)
&=
\frac{1}{2}
D_{KL}
\left(
P,
\frac{N_P P + N_Q Q}{N_P + N_Q}
\right)
\\
&\quad+
\frac{1}{2}
D_{KL}
\left(
Q,
\frac{N_P P + N_Q Q}{N_P + N_Q}
\right).
\end{aligned}
\]

After each scenario is added to a model—or when a new context is created—the agent computes the semi-weighted Jensen-Shannon divergence between all pairs of models associated with the corresponding action. If any pair of models $R_{C_j}^A$ and $R_{C_k}^A$ is found to be sufficiently similar, as determined by a divergence below the merging threshold $\Delta_m$, the agent proceeds to merge them: 

$$D_{swJS}(R_{C_j}^A, R_{C_k}^A) < \Delta_{m}$$

When two models are merged, one is replaced by the new merged model, while the other is discarded. All observations previously associated with the discarded model are reassigned to the merged model. This operation reduces the total number of models for the action by one and helps prevent fragmentation of the agent's representation space.

\subsection{Parameter Search for the Probabilistic Context Learner}
\label{sec : parameter_search}
The primary challenge for the Probabilistic MBRL algorithm lies in determining the optimal values for the thresholds. To assess the agent's performance, we have developed several metrics.

\begin{figure}[t]
\centering
\includegraphics[width=\columnwidth]{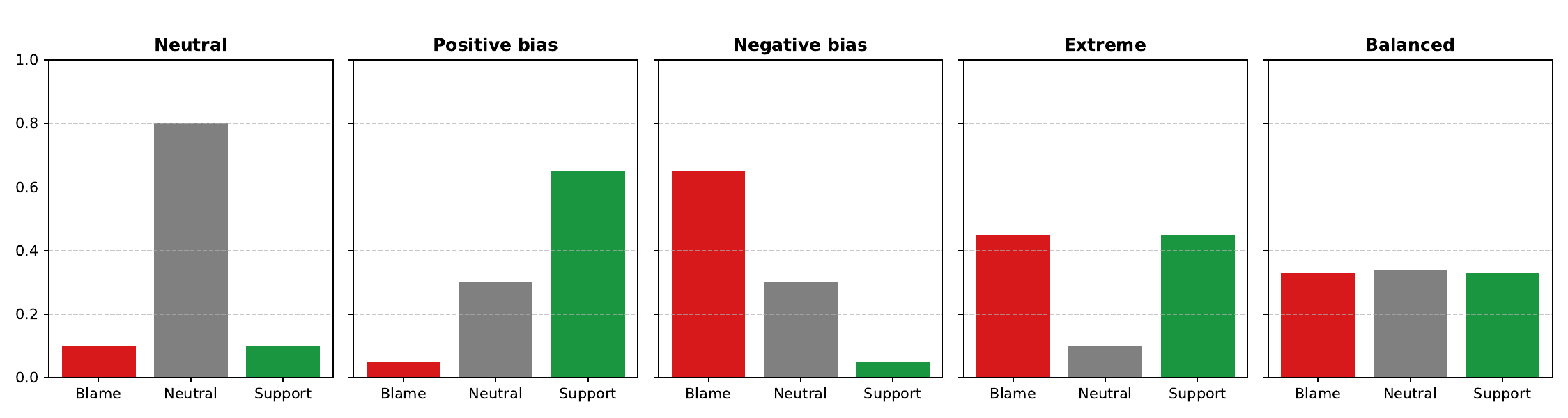}
\caption{\textbf{Ternary canonical distributions.} Figure shows the canonical distributions we created as a control for the Probabilistic Context Learner. They where built to represent the main possible types of ternary distributions we expect to encounter with the scenarios judgment distributions.}
\label{figA.1}
\end{figure}

Performance evaluation requires a set of distributions as inputs. To facilitate this, we constructed five canonical distributions from which we extract samples (Figure \ref{figA.1}). The testing set consists of samples drawn from each of these five canonical distributions. For each distribution, we collected $N=30$ samples of size $S=1000$. To ensure that the samples adhere to the canonical distributions, we set $S=1000$. This allows us to verify that the contexts formed are consistent with the corresponding samples.

Each testing sample is represented as a dictionary, analogous to the scenarios, with the keys \textit{action = 'test'}, \textit{reward = 'the sample'}, and \textit{scenario = 'the name of the canonical distribution from which the sample was drawn'}.

In total, we defined four distinct metrics to evaluate the algorithm’s performance as a function of the thresholds.

\subsubsection{Number of Contexts}

The first metric we defined is the number of contexts created. Since we know that in the ideal case, the algorithm creates 5 contexts, it allows us to evaluate if we obtain the right number of these. 

\subsubsection{Penalized Earth-Mover Distance}

The Earth-Mover Distance (EMD), also referred to as the Wasserstein Distance, quantifies the minimal cost necessary to transform one probability distribution into another. This metric is especially valuable for comparing distributions over categorical or ordinal spaces, such as a ternary distribution with outcomes in $\{-1,0,1\}$.

Given two probability distributions $P=(p_{-1}, p_0, p_1)$ and $Q=(q_{-1}, q_0, q_1)$ over the discrete set $\{-1,0,1\}$, the EMD is defined as:

$$ EMD = \sum_{i=-1}^{1} |F_P(i)-F_Q(i)|$$

Where $F_P$ and $F_Q$ are the cumulative distribution functions (CDFs) of $P$ and $Q$:

\begin{align*}
    &F_P(i)=\sum_{k \leq i}p_k,& F_Q(i)=\sum_{k \leq i}q_k
\end{align*}

The sum of absolute differences gives the total mass movement required to match $P$ to $Q$. We then applied the Earth-Mover Distance (EMD) to compare the distributions of the contexts obtained with the canonical distributions.

However, since the EMD is defined between two distributions, we employed a Hungarian-like algorithm to determine which distance to consider for each context distribution. Specifically, for each context distribution, we compute the EMD with all five canonical distributions, and select the minimal value as the corresponding EMD for that context. In the ideal scenario where the context distributions perfectly align with the canonical distributions, the EMD between our contexts and the canonical distributions would be zero.

Nevertheless, there are cases where this method may not be sufficient. This occurs when two distributions are similarly close to the same canonical distribution. Such a situation can arise if the distributions between the contexts are too similar or if more than five contexts are created. To account for these possibilities, we introduced a penalty, defined as:

\[
\mathrm{penalty}
=
\lambda \left(
\sum_{i=1}^{5}
\mathrm{Match}_i - 1
\right)
\]

We chose to fix the value of $\lambda$ of the same magnitude than the biggest EMD between canonical distributions which means that $\lambda=0.6$.

$$ EMD_{penalized} = EMD +\lambda$$

\subsubsection{Homogeneity of the contexts}

The third metric evaluates the homogeneity of the contexts. To calculate the homogeneity of a context, we first identify the canonical distribution that is most frequently represented among the samples within the context. We then calculate the ratio of this dominant canonical distribution to the total number of samples explained by the context. This yields the probability of encountering the dominant canonical distribution when randomly selecting a sample from within the context.

A homogeneity value closer to 1 indicates greater homogeneity within the context, signifying that the context predominantly reflects a single canonical distribution.

\subsubsection{Loss function}

Then we defined a loss function which we want to minimize. This loss function is defined thanks to the three metrics we presented above as: 

\[
\mathrm{Loss}
=
\mathrm{EMD}_{\mathrm{penalized}}
+
\lambda \left| N_{\mathrm{contexts}} - 5 \right|
+
\lambda \left| 1 - \frac{1}{\mathrm{Homogeneity}} \right|
\]

\subsubsection{Parametrization}

\begin{figure}[t]
\centering
\includegraphics[width=\columnwidth]{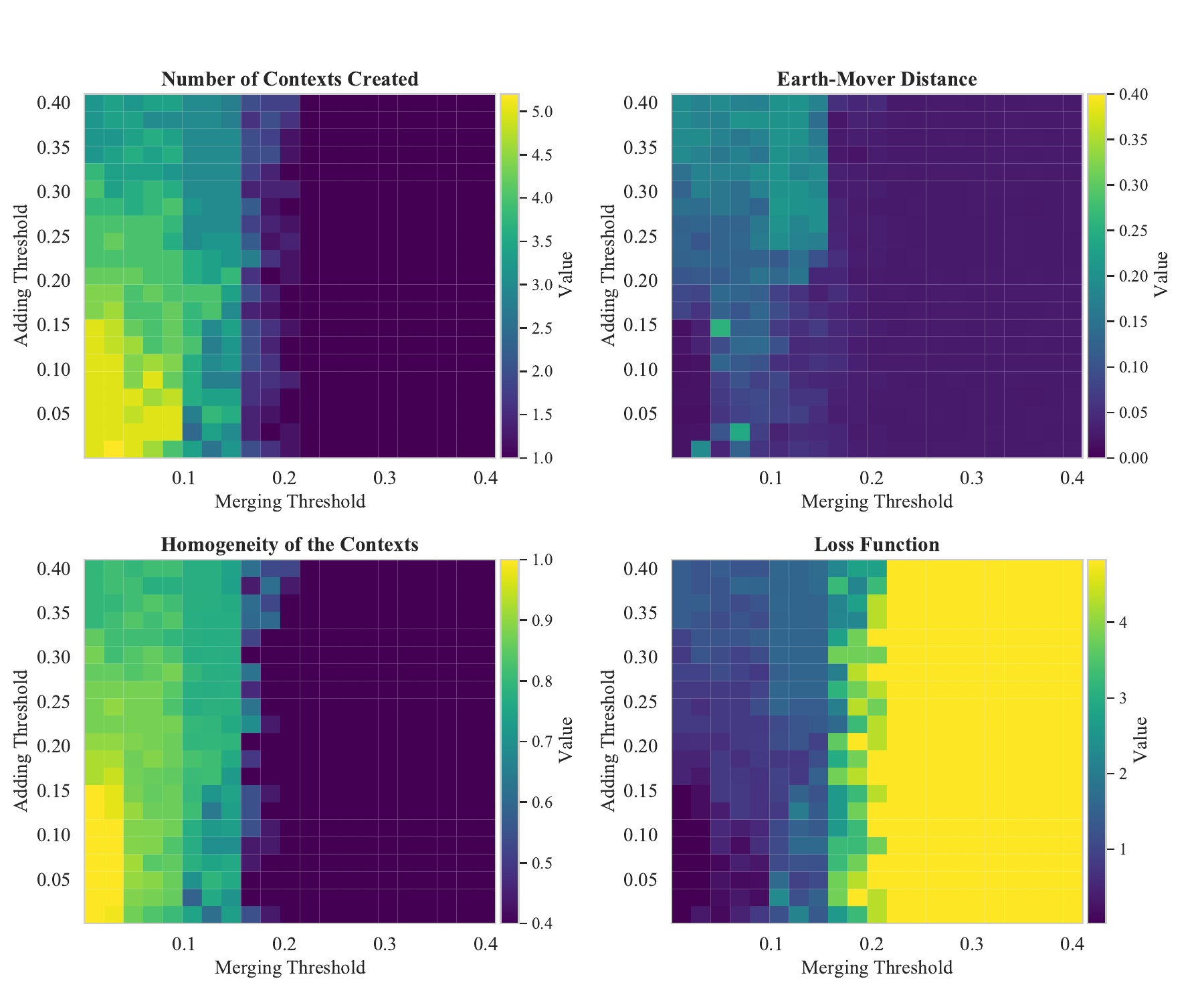}
\caption{\textbf{Metrics across threshold values} Heatmaps illustrating the threshold parameter search conducted for the MBRL agent. Each panel shows the mean over 10 runs for each threshold pair. Top-left: number of generated contexts, with an expected value of approximately 5. The top-right panel displays the penalized Earth Mover’s Distance (as defined in Section A.2), where lower values indicate better performance. Bottom-left: cluster homogeneity, ideally near 1. Bottom-right: overall loss combining previous metrics (Section A.4).}
\label{fig:quadr_test_propre}
\end{figure}

\begin{figure}[t]
\centering
\includegraphics[width=\columnwidth]{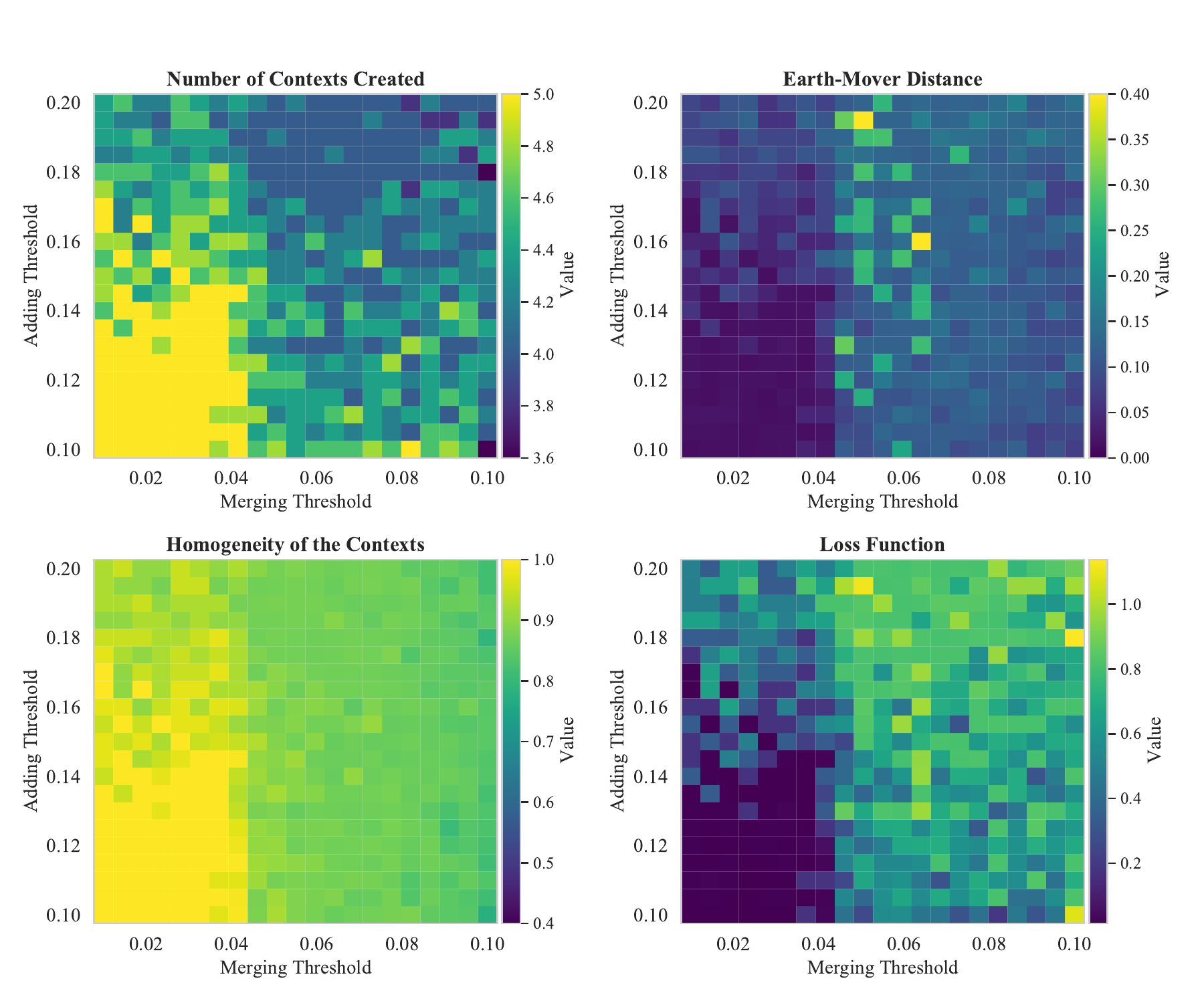}
\caption{\textbf{Metrics across threshold values with a finer interval.} Heatmaps illustrating the threshold parameter search conducted for the MBRL agent. Each panel shows the mean over 10 runs for each threshold pair. Top-left: number of generated contexts, with an expected value of approximately 5. Top-right: penalized Earth Mover’s Distance (Section A.2), where lower values indicate better performance. Bottom-left: cluster homogeneity, ideally near 1. Bottom-right: overall loss combining previous metrics (Section A.4). The optimal threshold range is $\Delta_m \in [0.005 ; 0.02]$ and $\Delta_a\in[0.10;0.14]$.}

\label{fig:quadr_test_sharper_propre}
\end{figure}

We considered threshold values of $\Delta_a \in [0.01,0.4]$ and $\Delta_a \in [0.01,0.4]$ with $N=30$. The results are calculated with a mean over five iterations, with random test datasets of size 150. 
Figure \ref{fig:quadr_test_propre} shows the results we obtained. Since the goal is to have a number of contexts equal to five, a penalized EMD and a loss function the smaller possible and an homogeneity the closer to 1, we can see that the optimal values for the thresholds are $\Delta_a \in [0.10,0.20]$ and $\Delta_m \in [0.01,0.10]$ (Figure \ref{fig:quadr_test_sharper_propre}).

We then considered smaller intervals for the threshold values (Figure \ref{fig:quadr_test_sharper_propre}). The ideal values for the thresholds are $\Delta_a = 0.12$ and $\Delta_m = 0.03$.

\subsubsection{Results with optimal thresholds}
\label{subsec : results_opt_th}
\begin{figure}[t]
    \centering

    \begin{minipage}{0.9\columnwidth}
        \centering
        \includegraphics[width=0.8\columnwidth]{figures/canonical_distributions.pdf}
        
        \small (a) Canonical distributions
    \end{minipage}

    \vspace{0.2cm}

    \begin{minipage}{0.9\columnwidth}
        \centering
        \includegraphics[width=0.8\columnwidth]{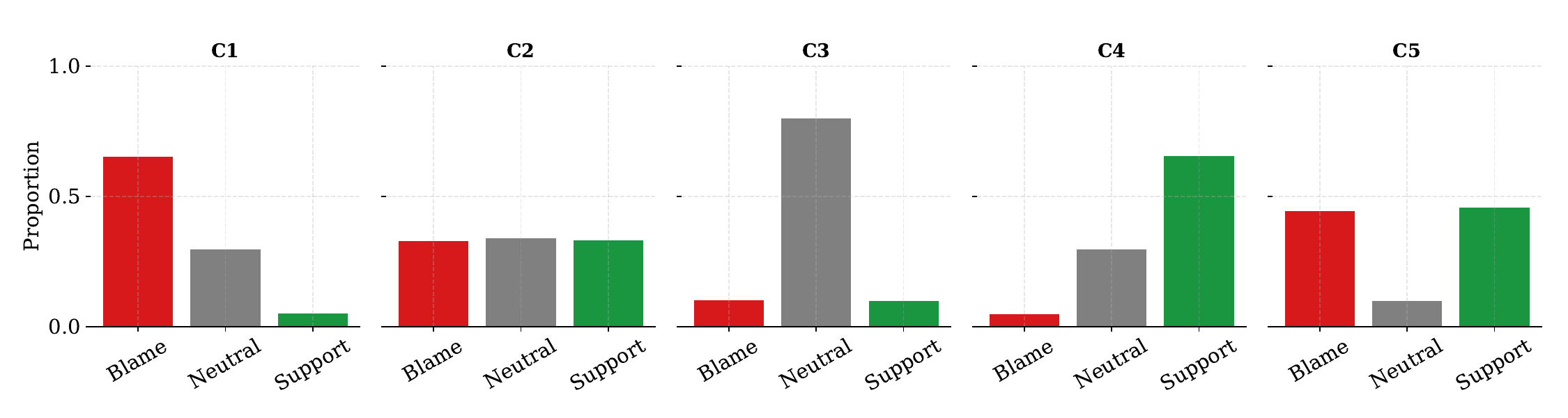}
        
        \small (b) Contexts obtained
    \end{minipage}

    \caption{\textbf{Comparison between the canonical distributions (top) and the contexts obtained (bottom).} The contexts are obtained with a merging threshold of $\Delta_m=0.03$ and an adding threshold of $\Delta_a=0.12$, the number of samples taken from each canonical distribution is $N=30$, and they are all of size $S=1000$. The Probabilistic Context Learner succeeds in creating contexts corresponding to the canonical distributions from which the samples were derived.}
    
    \label{fig:figures_superposees}
\end{figure}

To assess whether these thresholds are appropriately established, we visualize the distributions resulting from their application. As shown in Figure \ref{fig:figures_superposees}, the contexts align perfectly with the canonical distributions. Furthermore, the homogeneity of these contexts is equal to 1, indicating an ideal match between the generated contexts and their corresponding canonical distributions.

\subsubsection{Control test}
\label{subsec : one_context}

\begin{figure}[t]
\centering
\includegraphics[width=\columnwidth]{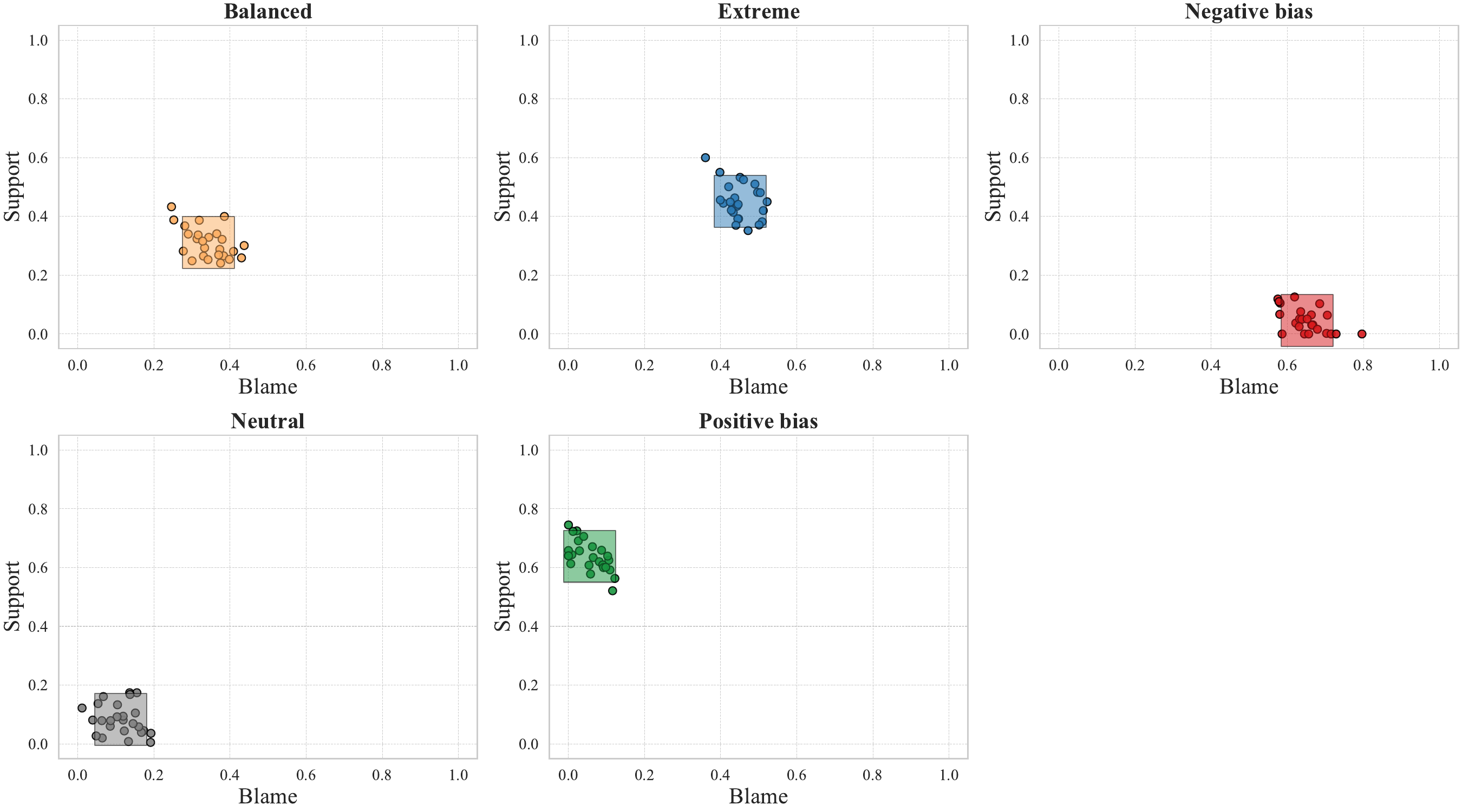}
\caption{\textbf{The Probabilistic Context Learner infers a single context when all samples originate from a common distribution.} Each subplot corresponds to a specific canonical distribution and displays the clustering performed by the agent. Points represent individual samples derived from the canonical distribution with a noise $\eta=0.10$, while squares denote the clusters (or contexts) created by the agent. The size of each square is proportional to the number of scenarios it contains. Coordinates reflect P(Support)P(Support) and P(Blame)P(Blame); the third outcome (Neutral) is implicitly defined as $1-P(Support)-P(Blame)$, allowing for a two-dimensional representation.}
\label{fig:control_noise}
\end{figure}

As a control test, we evaluate whether the Probalistic MBRL agent creates only one context when the dataset is composed of scenarios with close distributions. We create noisy samples from one canonical distribution. A noise level from a uniform distribution of size $\eta$ is added to the canonical distributions value. The distributions are re-normalized to keep the probabilistic nature of the distributions.

On Figure \ref{fig:control_noise} the noise level is fixed to $\eta = 0.10$, which represents one third of the \textit{Balanced} distribution value. Figure \ref{fig:control_noise} shows that when the samples follow similar distributions, the MBRL agent creates one context only, for all canonical distributions.

\subsubsection{Sensitivity to the Hyper-parameters}

\begin{figure*}[!t]
    \centering
    \includegraphics[width=0.95\textwidth]{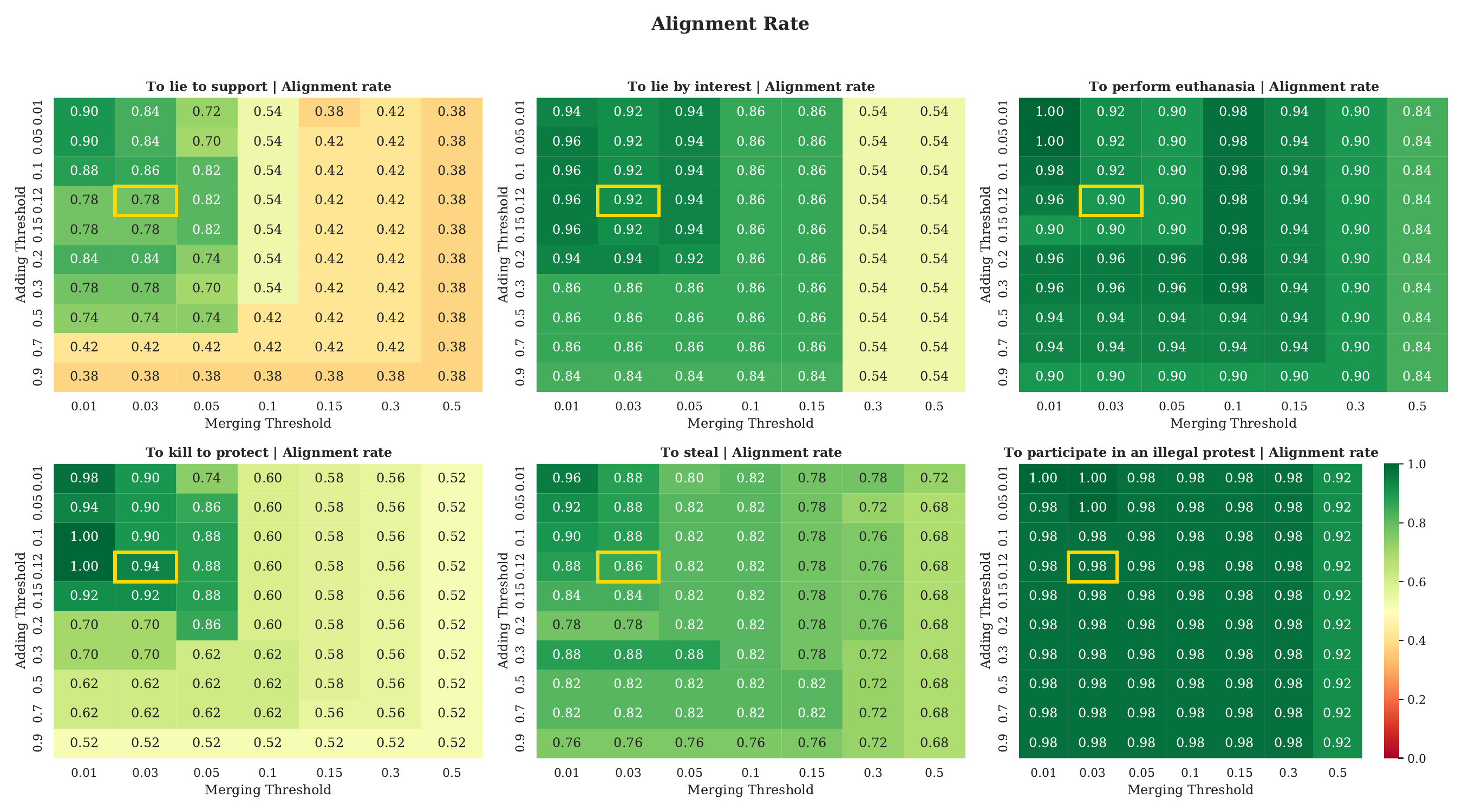}
    \caption{\textbf{Alignment rate for different pairs of thresholds for each action.} The golden square indicates the threshold values used in the main experiments ($\Delta_a$=0.12, $\Delta_m$=0.03)}
    \label{fig:alig_rate_th}
\end{figure*}

\begin{figure*}[!t]
    \centering
    \includegraphics[width=0.95\textwidth]{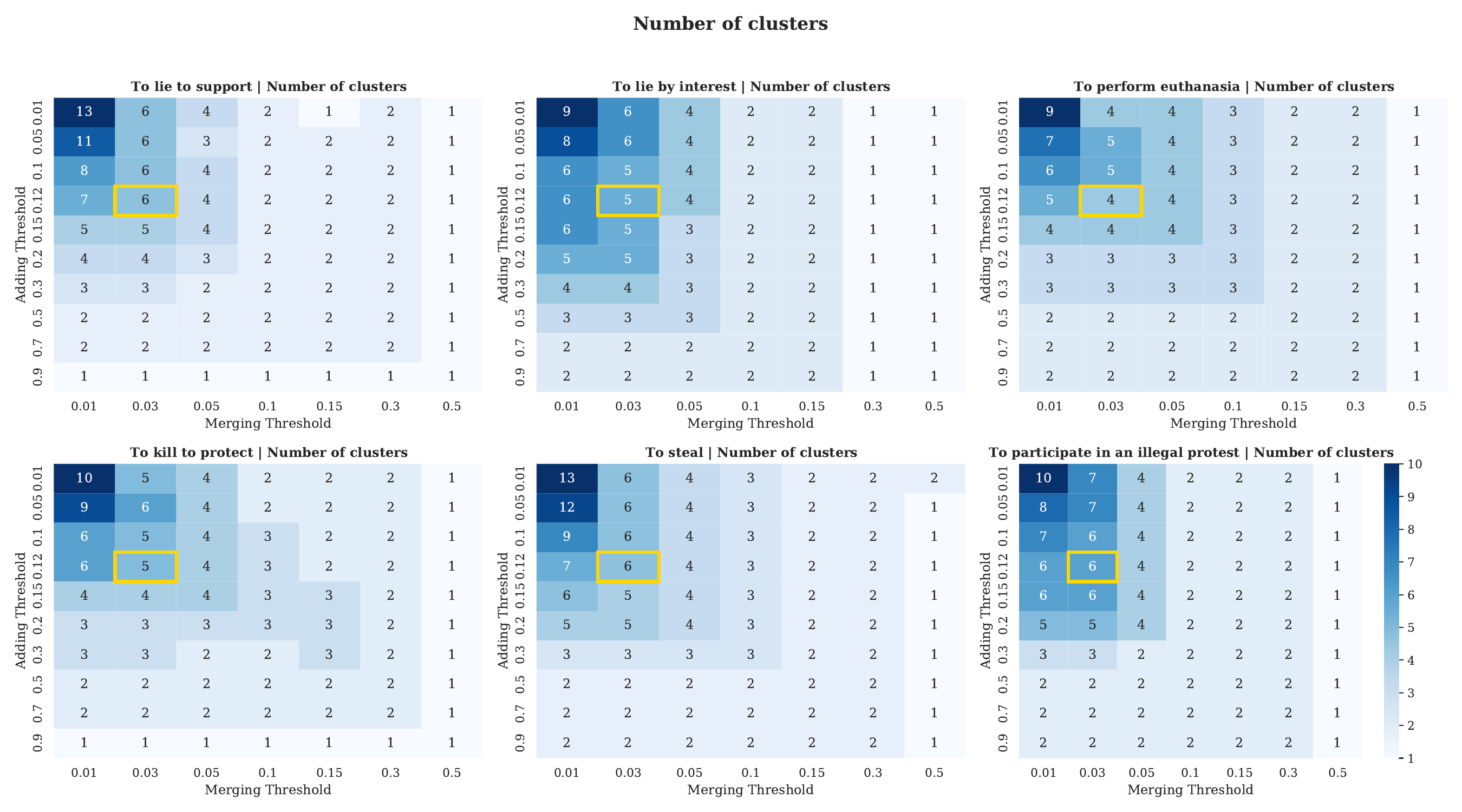}
    \caption{\textbf{Number of clusters for different pairs of thresholds for each action.} The golden square indicates the threshold values used in the main experiments ($\Delta_a$=0.12, $\Delta_m$=0.03)}
    \label{fig:nb_clusters_th}
\end{figure*}

\begin{figure*}[!t]
    \centering
    \includegraphics[width=0.95\textwidth]{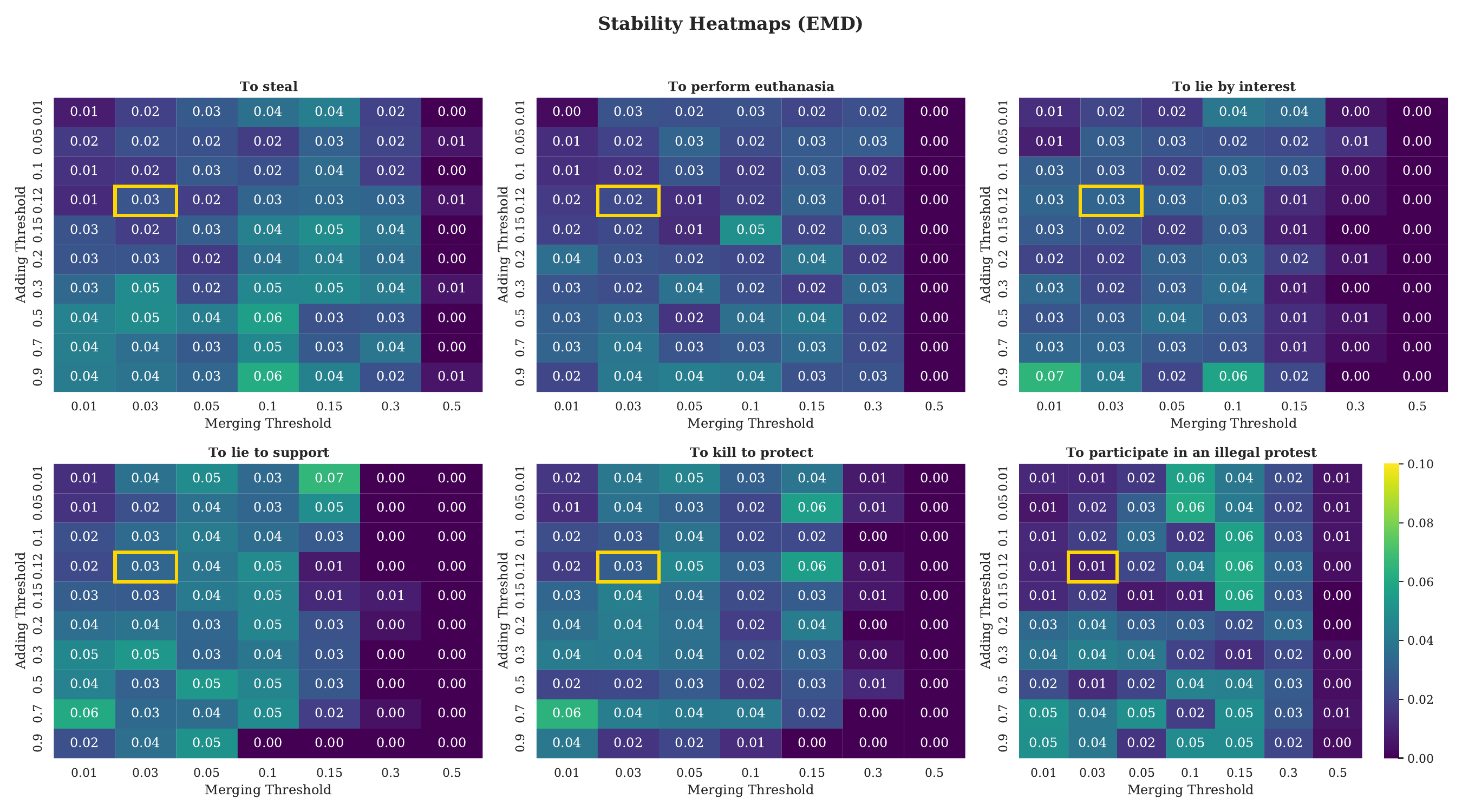}
    \caption{\textbf{Stability for different pars of thresholds for each action.} Stability is calculated by making 5 clusterings with data arriving in a randomize order at each time. We then compare the Earth Mover Distance between matching clusters (thanks to an Hungarian algorithm finding the nearest clusters). It gives a matrix 5x5 from which we take the mean to have the value for this pair of threshold. A score below 0.10 means that the clusters do not variate so much between seeds.}
    \label{fig:stab_th}
\end{figure*}

We evaluated the sensitivity of our model to the threshold parameters. To address this, we conducted a short study varying both the adding threshold ($\Delta_a$) and the merging threshold ($\Delta_m$). We measured three metrics: alignment rate (Figure \ref{fig:alig_rate_th}), number of clusters (Figure \ref{fig:nb_clusters_th}), and robustness (Figure \ref{fig:stab_th}).

The results are summarized in the accompanying figures, where the golden square indicates the threshold values used in the main experiments ($\Delta_a$=0.12, $\Delta_m$=0.03). These values represent the best compromise between maximizing alignment rate and maintaining a manageable number of clusters. Importantly, the model demonstrates high robustness across all tested threshold pairs, indicating that our clustering procedure is stable and not overly sensitive to precise threshold settings.

\subsection{Pre-processing}
\label{sec: app_pre-processing}
The original formulations of the moral scenarios could not be directly processed by the Probabilistic Context Learner, which requires consistent and standardized representations of the underlying moral actions. Without such standardization, the agent would be forced to cluster scenarios solely based on similarities in human judgment distributions, risking the conflation of semantically distinct actions (e.g., “to lie to a child” vs. “to perform euthanasia”) and failing to recognize semantically equivalent actions expressed differently (e.g., “to perform euthanasia” vs. “to inject lethal drugs”). This distinction is crucial: actions that appear superficially similar in judgment profiles may correspond to fundamentally different moral rules, while different wordings of the same action must be recognized as equivalent to support meaningful generalization.

To address this,\textsc{COMETH} incorporates a dedicated pre-processing pipeline, with the first stage referred to as the LLM filter. In this stage, we employ a few-shot prompting strategy using open-source large language models (Mistral-7B-Instruct-v0.3, Llama-3.1-8B-Instruct, and Qwen-3-Next-80B-A3B-Instruct; \citealp{qwen2.5-1m}) to generate abstracted representations of each scenario. The LLM is prompted to identify the principal verb of the action and condense both the subject and the complement, producing a concise, semantically uniform expression of the form: \emph{“To + verb + summary of the complement”}.

This transformation isolates the core action while stripping away extraneous contextual details, thereby enabling semantically equivalent actions across different scenarios to be treated as the same abstract action and distinct actions to remain separate. For example, “to inject lethal drugs” and “to perform euthanasia” would be mapped to the same abstract representation, while “to lie about Santa Claus” remains distinct from “to steal food for a starving child”, despite potentially similar moral judgment profiles.

After applying the LLM filter, each generalized action is embedded using the all-MiniLM-L6-v2 Sentence Transformer, which produces a numerical representation suitable for clustering. A K-means algorithm is then applied to these embeddings, grouping scenarios into clusters that represent the Core Actions. These Core Actions serve as the fundamental units in the \textsc{COMETH}{COMETH} pipeline, allowing subsequent stages—such as moral evaluation generalization and context modeling—to operate on standardized and comparable action representations.

This pipeline ensures that the Probabilistic Context Learner learns associations between human moral judgments and semantically meaningful actions rather than superficial textual similarities.

\subsubsection{Pre-Processing Prompts}
\label{sec:pre_proc_prompts}
We used five prompt strategies to extract the core moral action from scenarios. All prompts are displayed in Table \ref{tab:preprocessing_prompts}.

\begin{table}[t]
\centering
\small
\begin{tabular}{p{0.95\columnwidth}}
\toprule
\multicolumn{1}{c}{\textbf{LLM Prompts used for the pre-processing}} \\
\midrule
\textbf{A\_Minimalist:} Extract the \textbf{main literal action} of the scenario in at most two words. Ignore targets, motivations, consequences, or locations --- keep only the observable action itself. \\
\addlinespace

\textbf{B\_Infinitive:} Summarize the scenario as a \textbf{single verb in the infinitive form} (e.g., ``to drive'', ``to cheat''). Ignore targets, motivations, consequences, or locations --- keep only the observable action itself. \\
\addlinespace

\textbf{C\_MainAct:} Identify the \textbf{main act performed} in the scenario. Answer with a short phrase describing the literal action. Ignore targets, motivations, consequences, or locations --- keep only the observable action itself. \\
\addlinespace

\textbf{D\_OneWord:} Provide a \textbf{one-word label} that best captures the observable action of the scenario. Ignore targets, motivations, consequences, or locations --- keep only the observable action itself. \\
\addlinespace

\textbf{E\_NounPhrase:} Summarize the scenario as a concise \textbf{noun phrase} describing the main literal action (e.g., ``contract killing'', ``bank robbery''). Ignore targets, motivations, consequences, or locations --- keep only the observable action itself. \\
\midrule
\textbf{Scenario:} [...] \\
\bottomrule
\end{tabular}
\caption{The five pre-processing prompt variations designed to extract the core action from the generated scenarios.}
\label{tab:preprocessing_prompts}
\end{table}

\subsection{Generalization}
\label{sec: app_generalization}

After clustering the scenarios according to human moral evaluations using the Probabilistic Context Learner, we address the questions of interpretability and predictive capacity through the introduction of a Generalization module.

This module leverages an LLM to infer descriptive contextual features that characterize each cluster. These features represent properties that are consistently shared among scenarios within the same context but absent from others, thereby offering candidate explanations for differences in moral judgment. Importantly, these are contextual rather than evaluative features; their extraction relies solely on semantic content and does not involve any moral labeling. Each feature is encoded as a concise binary statement (e.g., “Consent expressed by the victim”), with a fixed number of features extracted per cluster.

From this feature set, we construct a collection of binary variables, which are systematically evaluated across all scenarios using the same LLM. This procedure assigns each scenario a binary feature vector that encodes the presence or absence of the extracted contextual features. At the cluster level, each context is represented by an aggregate feature profile, computed as the frequency of each feature across all scenarios belonging to that context, weighted by the relative frequency of the context in the dataset.

To quantify the relationship between scenarios and contexts, we compute a similarity score between each scenario and each context using a log-likelihood function. The function is weighted by learnable feature importance parameters, which capture the contribution of each feature. Scores are normalized with a softmax function to obtain a probability distribution over contexts. Model training proceeds by minimizing the negative log-likelihood of the true context labels, with L2 regularization to reduce overfitting. Optimization is performed using the L-BFGS-B algorithm, and model performance is assessed via 25-fold cross-validation, where each fold contains two held-out scenarios. This procedure yields interpretable weights associated with individual features, directly linking them to predictive performance.

\subsubsection{Benchmark Evaluation}

To validate the proposed approach, we conducted a benchmark on a “To Steal” dataset (50 scenarios), different than the one presented in the results of this study  with features extraction and evaluation using OpenAI GPT-4. We compared our negative log-likelihood (NLL) model against standard classifiers, using 25-fold cross-validation to predict cluster membership. Results showed that Random Forest achieved 40\% accuracy, Support Vector Machine (SVM) 54\%, Logistic Regression 46\%, while our NLL model reached 82\% accuracy, significantly outperforming all baselines.

In addition to predictive performance, the NLL model offers enhanced interpretability compared to black-box methods. Each scenario is represented as a binary vector indicating the presence or absence of explicitly defined contextual features (e.g., “The agent had no alternative,” “The victim gave consent”). The model learns an importance weight for each feature, quantifying its contribution to the probability of assigning a scenario to a given moral context. These weights are directly accessible and interpretable, providing clear insights into which contextual factors most strongly influence classification outcomes.

By contrast, black-box models such as Random Forests or deep neural networks obscure the link between input features and output predictions. Our approach, in contrast, offers a transparent mapping from human-understandable features to model decisions, thereby enabling users to trace and understand the reasoning process. This fulfills the interpretability requirement of the \textsc{COMETH}{COMETH} pipeline, making the model suitable both for scientific analysis and for deployment in morally sensitive applications.

\subsubsection{Feature Extraction Prompts}

This subsection presents the three prompts we used in order to extract the features. Each prompt was submitted to the three open source LLMs we used, namely mistralai/Mistral-7B-Instruct-v0.3, meta-llama/Llama-3.1-8B-Instruct and Qwen/Qwen3-Next-80B-A3B-Instruct (\citealp{qwen2.5-1m}). In this method, the action are considered independently. Then, the clusters we present to the LLMs are the clusters of scenarios created by the Probabilistic Context Learner. The different prompts are shown in Table \ref{tab:feature_extraction_prompt1}, \ref{tab:feature_extraction_prompt2} and \ref{tab:feature_extraction_prompt3}.

\begin{table}[t]
\centering
\small
\begin{tabular}{p{0.95\columnwidth}}
\toprule
\multicolumn{1}{c}{\textbf{Feature Extraction Prompt 1}} \\
\midrule
You are given 5 clusters of scenarios. Each cluster contains short scenarios describing the same action performed in different contexts. \\
\addlinespace
Your task is to generate 5 short contextual features for each cluster. \\
\midrule
\textbf{Constraints:} \\
$\bullet$ Features must be descriptive only, without moral or evaluative terms (avoid words like good, bad, justified, unfair). \\
$\bullet$ Within a cluster, each feature must be shared by multiple scenarios in that cluster. \\
$\bullet$ Features should be distinctive across clusters: avoid features that could equally describe scenarios in other clusters. \\
$\bullet$ Write features as short noun phrases (2--5 words), focusing on observable elements such as participants, environment, tools, or conditions. \\
$\bullet$ Provide exactly 5 features per cluster. \\
\midrule
\textbf{Input (5 clusters of scenarios):} \\
Cluster 1: [Insert scenarios] \\
Cluster 2: [Insert scenarios] \\
Cluster 3: [Insert scenarios] \\
Cluster 4: [Insert scenarios] \\
Cluster 5: [Insert scenarios] \\
\midrule
\textbf{Output (5 contextual features per cluster):} \\
Cluster 1: ... \\
Cluster 2: ... \\
(and so on for all 5 clusters) \\
\bottomrule
\end{tabular}
\caption{Zero-shot feature extraction prompt defining strict structural and semantic constraints.}
\label{tab:feature_extraction_prompt1}
\end{table}

\begin{table}[t]
\centering
\small
\begin{tabular}{p{0.95\columnwidth}}
\toprule
\multicolumn{1}{c}{\textbf{Feature Extraction Prompt 2}} \\
\midrule
You are given 5 clusters of scenarios. Each cluster contains about 50 short scenarios describing the same action performed in different contexts. \\
\addlinespace
Your task is to identify 5 short contextual features for each cluster. To ensure that features are robust and distinctive, follow this two-step process for each cluster: \\
\addlinespace
$\bullet$ \textbf{Candidate Extraction:} List all contextual elements that appear repeatedly across scenarios in the cluster (e.g., armed aggressor, public place, presence of children). Exclude moral or evaluative terms. \\
$\bullet$ \textbf{Feature Selection:} From the candidate list, choose the 5 most frequent and distinctive features that are: \\
\quad -- shared by multiple scenarios in this cluster, and \\
\quad -- not generally present in other clusters of the same action. \\
\midrule
\textbf{Guidelines:} \\
$\bullet$ Write features as short noun phrases (2--5 words). \\
$\bullet$ Avoid redundancy. \\
$\bullet$ Focus on observable context elements such as participants, environment, tools, or conditions. \\
$\bullet$ Provide exactly 5 features per cluster. \\
\midrule
\textbf{Input (5 clusters of scenarios):} \\
Cluster 1: [Insert scenarios] \\
Cluster 2: [Insert scenarios] \\
Cluster 3: [Insert scenarios] \\
Cluster 4: [Insert scenarios] \\
Cluster 5: [Insert scenarios] \\
\midrule
\textbf{Output (5 contextual features per cluster):} \\
Cluster 1: \\
\quad Candidate features: ... \\
\quad Selected features: ... \\
Cluster 2: \\
\quad Candidate features: ... \\
\quad Selected features: ... \\
(and so on for all 5 clusters) \\
\bottomrule
\end{tabular}
\caption{Feature extraction prompt incorporating a two-step reasoning process (Candidate Extraction and Feature Selection).}
\label{tab:feature_extraction_prompt2}
\end{table}

\begin{table}[t]
\centering
\small
\begin{tabular}{p{0.95\columnwidth}}
\toprule
\multicolumn{1}{c}{\textbf{Feature Extraction Prompt 3}} \\
\midrule
\textbf{Task.} You are given 5 clusters of scenarios describing the same action in different contexts. Extract 5 contextual features per cluster. \\
\midrule
\textbf{Rules.} Features must be contextual (no moral terms). Each feature must appear in multiple scenarios within a cluster. Features must be distinctive across clusters. Use short noun phrases (2--5 words). Output exactly 5 features per cluster. \\
\midrule
\textbf{Example (toy case).} \\
\addlinespace
\textbf{Cluster 1 (survival theft):} \\
Grocery store setting; Food theft; Survival motivation; Low-income context; Market environment. \\
\addlinespace
\textbf{Cluster 2 (luxury theft):} \\
Luxury goods target; Mall environment; Concealment behavior; High-value items; Boutique setting. \\
\midrule
\textbf{Task.} Apply the same method to: \\
\addlinespace
Cluster 1: [...] \\
Cluster 2: [...] \\
Cluster 3: [...] \\
Cluster 4: [...] \\
Cluster 5: [...] \\
\midrule
\textbf{Output.} Exactly 5 features per cluster, formatted as lists of short phrases. \\
\bottomrule
\end{tabular}
\caption{The few-shot prompt used to extract contextual features from clusters of scenarios.}
\label{tab:feature_extraction_prompt3}
\end{table}

\section{Extended Results}
In this section, we provide a detailed overview of the results obtained across the full COMETH pipeline, complementing the summary figures presented in the main text. We report comprehensive evaluations of the pre-processing (see the Extended Pre-processing Results Section), end-to-end LLMs (Extended End-to-end LLMs Results section), and generalization modules (Extended Generalization Results section), including comparisons across multiple LLMs, prompts, and moral actions. These extended results highlight the robustness, consistency, and interpretability of our approach, and allow for a deeper understanding of how different design choices—such as model size, prompt formulation, and feature representation—affect alignment with human moral judgments. All metrics presented here are intended to provide transparency and reproducibility.

\subsection{Extended Pre-processing Results}
\label{sec:app_res_pre-processing}

We evaluated three LLMs (Mistral 7B, Qwen 7B, Qwen 80B) with five prompting strategies using ARI, NMI, and V-measure to assess how well scenario representations recover context clusters. As shown on Table \ref{tab:clustering_metrics} Qwen 80B consistently achieves the highest scores, indicating that larger models produce more robust action representations. Minimalist and MainAct prompts generally perform best across models, while OneWord prompts underperform. NounPhrase prompts are especially effective for Mistral 7B. The alignment of ARI, NMI, and V-measure suggests that accurate clustering also preserves information-theoretic structure. These results indicate that larger models combined with concise, informative prompts yield the most reliable clusters of moral scenarios, supporting downstream analysis in \textsc{COMETH}.

\begin{table}[t]
\centering
\begin{tabular}{|c|c|c|c|c|}
\hline
\textbf{LLM} & \textbf{Prompt} & \textbf{ARI} & \textbf{NMI} & \textbf{V-measure} \\
\hline
\multirow{5}{*}{Mistral 7B} 
& Minimalist & 0.65 & 0.74 & 0.74 \\
& Infinitive & 0.53 & 0.63 & 0.63 \\
& MainAct & 0.83 & 0.85 & 0.85 \\
& OneWord & 0.69 & 0.77 & 0.77 \\
& NounPhrase & 0.89 & 0.89 & 0.89 \\
\hline
\multirow{5}{*}{Llama 8B} 
& Minimalist & 0.78 & 0.81 & 0.81 \\
& Infinitive & 0.54 & 0.65 & 0.65 \\
& MainAct & 0.79 & 0.82 & 0.82 \\
& OneWord & 0.43 & 0.60 & 0.60 \\
& NounPhrase & 0.59 & 0.72 & 0.72 \\
\hline
\multirow{5}{*}{Qwen 80B} 
& Minimalist & 0.90 & 0.91 & 0.91 \\
& Infinitive & 0.68 & 0.76 & 0.76 \\
& MainAct & 0.87 & 0.88 & 0.88 \\
& OneWord & 0.59 & 0.72 & 0.72 \\
& NounPhrase & 0.87 & 0.89 & 0.89 \\
\hline
\end{tabular}
\caption{Clustering performance (ARI, NMI, V-measure) for each LLM and prompt.}
\label{tab:clustering_metrics}
\end{table}

These extended results confirm that the pre-processing step in \textsc{COMETH} reliably produces semantically coherent clusters across LLMs and prompts. Qwen 80B consistently achieves the highest ARI, NMI, and V-measure scores, reflecting robust and stable action representations. Minimalist and MainAct prompts perform well across models, whereas OneWord prompts generally underperform. Notably, NounPhrase prompts are particularly effective for Mistral 7B. The consistent alignment between ARI, NMI, and V-measure indicates that the clustering preserves both label consistency and information-theoretic structure. Overall, these results support the reproducibility and reliability of the LLM-based pre-processing pipeline, providing a solid foundation for subsequent generalization and moral evaluation tasks.
 
\subsection{Extended End-to-end LLMs Results}
\label{sec:detailed_end_to_end}

\subsubsection{End-to-End Moral Evaluation Prompts}
\label{sec:end_to_end_prompts}

We used five prompt strategies to elicit moral judgments from LLMs. All prompts are available on Table \ref{tab:llm_prompts}.

\begin{table}[t]
\centering
\small 
\begin{tabular}{p{0.95\columnwidth}}
\toprule
\multicolumn{1}{c}{\textbf{LLM Moral Judgment Prompts}} \\
\midrule
\textbf{Prompt 1:} Read the scenario and judge the morality of the main action. Answer only with one of these options: ``Support'', ``Blame'', or ``Neutral''. \\
\addlinespace 

\textbf{Prompt 2:} Evaluate the scenario and determine whether the action is morally acceptable, wrong, or neutral. Respond only with ``Support'', ``Blame'', or ``Neutral''. \\
\addlinespace

\textbf{Prompt 3:} Assess the scenario and provide a moral judgment: ``Support'', ``Blame'', or ``Neutral''. Also give a confidence interval (0–100) expressing how certain you are about your judgment. \\
\addlinespace

\textbf{Prompt 4:} Judge the morality of the scenario's main action. Answer ``Support'', ``Blame'', or ``Neutral''. Include a confidence score between 0 and 100 reflecting your certainty. \\
\addlinespace

\textbf{Prompt 5:} Consider the scenario and provide a concise moral judgment. Use only ``Support'', ``Blame'', or ``Neutral''. \\
\midrule
\textbf{Scenario:} [...] \\
\bottomrule
\end{tabular}
\caption{The five prompt variations used to elicit moral judgments from the LLMs, followed by the scenario placeholder.}
\label{tab:llm_prompts}
\end{table}

\subsubsection{End-to-end LLMs results}

In this section, we report the detailed results obtained with end-to-end LLM approaches. Table~\ref{tab:llm_alig_rate} presents the alignment rates of each model across six morally relevant actions and five different prompts. Overall, the results indicate that end-to-end approaches exhibit limited and variable capacity to reproduce human moral judgments. Llama 8B, for instance, shows high error rates -defined as the fact of not returning a ternary evaluation as asked by the prompt- and inconsistencies across prompts (Table \ref{tab:error_rates}), reducing its reliability as a standalone moral evaluator. Mistral 7B performs moderately, achieving reasonable alignment on some actions but showing variability across prompts and actions.

By contrast, Qwen 80B demonstrates robust adherence to the requested response format (‘Support’, ‘Blame’, or ‘Neutral’) and shows consistent behavior across prompts. While the mean alignment rates remain modest (generally 0.44–0.46), Qwen 80B is markedly more stable than the other models, highlighting that larger and more capable LLMs can produce more coherent moral judgments even without structured feature representations. These findings emphasize the limitations of black-box, end-to-end moral evaluation, particularly for smaller or less instruction-tuned models, and underscore the need for structured pipelines—such as \textsc{COMETH}—to significantly improve both accuracy and consistency.

In sum, while end-to-end LLMs can capture some aspects of human judgment, their variable performance and prompt sensitivity highlight the importance of combining them with structured pre-processing and feature-based reasoning to achieve reliable, generalizable moral alignment.

\begin{table*}[t]
\centering
\begin{tabular}{llccccccc}
\toprule
\textbf{LLM} & \textbf{Prompt} & \textbf{Lie by Interest} & \textbf{Lie to Support} & \textbf{Euthanasia} & \textbf{Kill to Protect} & \textbf{Steal} & \textbf{Protest} & \textbf{Mean} \\
\midrule
\multirow{5}{*}{Mistral 7B} 
& Prompt 1 & 0.18 & 0.27 & 0.54 & 0.23 & 0.42 & 0.49 & 0.36 \\
& Prompt 2 & 0.19 & 0.24 & 0.46 & 0.21 & 0.33 & 0.40 & 0.31 \\
& Prompt 3 & 0.28 & 0.32 & 0.59 & 0.31 & 0.41 & 0.40 & 0.38 \\
& Prompt 4 & 0.24 & 0.31 & 0.75 & 0.37 & 0.48 & 0.39 & 0.41 \\
& Prompt 5 & 0.26 & 0.25 & 0.47 & 0.22 & 0.39 & 0.46 & 0.34 \\
\midrule
\multirow{5}{*}{Llama 8B} 
& Prompt 1 & 0.27 & 0.27 & 0.98 & 0.49 & 0.38 & 0.43 & 0.47 \\
& Prompt 2 & 0.21 & 0.23 & 0.94 & 0.46 & 0.33 & 0.33 & 0.42 \\
& Prompt 3 & 0.34 & 0.54 & 0.98 & 0.86 & 0.45 & 0.66 & 0.64 \\
& Prompt 4 & 0.27 & 0.44 & 1.00 & 0.89 & 0.42 & 0.57 & 0.60 \\
& Prompt 5 & 0.14 & 0.28 & 0.85 & 0.40 & 0.28 & 0.50 & 0.41 \\
\midrule
\multirow{5}{*}{Qwen 80B} 
 & Prompt 1 & 0.42 & 0.58 & 0.24 & 0.54 & 0.36 & 0.64 & 0.46 \\
 & Prompt 2 & 0.44 & 0.60 & 0.10 & 0.54 & 0.34 & 0.60 & 0.44 \\
 & Prompt 3 & 0.44 & 0.60 & 0.10 & 0.54 & 0.36 & 0.68 & 0.45 \\
 & Prompt 4 & 0.46 & 0.56 & 0.18 & 0.54 & 0.30 & 0.58 & 0.44 \\
 & Prompt 5 & 0.46 & 0.60 & 0.14 & 0.54 & 0.36 & 0.60 & 0.45 \\
\bottomrule
\end{tabular}
\caption{Alignment rate of each End-to-end LLM on each action for the five prompts we used.}
\label{tab:llm_alig_rate}
\end{table*}

\begin{table*}[t]
\centering
\begin{tabular}{llccccccc}
\toprule
\textbf{LLM} & \textbf{Prompt} & \textbf{Lie by Interest} & \textbf{Lie to Support} & \textbf{Euthanasia} & \textbf{Kill to Protect} & \textbf{Steal} & \textbf{Protest} & \textbf{Mean} \\
\midrule
\multirow{5}{*}{Mistral 7B} 
& Prompt 1 & 0.00 & 0.00 & 0.02 & 0.02 & 0.00 & 0.00 & 0.01 \\
& Prompt 2 & 0.00 & 0.00 & 0.02 & 0.02 & 0.00 & 0.00 & 0.01 \\
& Prompt 3 & 0.12 & 0.06 & 0.44 & 0.18 & 0.16 & 0.20 & 0.19 \\
& Prompt 4 & 0.04 & 0.08 & 0.42 & 0.22 & 0.14 & 0.16 & 0.18 \\
& Prompt 5 & 0.00 & 0.02 & 0.00 & 0.00 & 0.00 & 0.02 & 0.01 \\
\midrule
\multirow{5}{*}{Llama 8B} 
& Prompt 1 & 0.00 & 0.08 & 0.94 & 0.38 & 0.00 & 0.16 & 0.26 \\
& Prompt 2 & 0.00 & 0.04 & 0.94 & 0.34 & 0.02 & 0.16 & 0.25 \\
& Prompt 3 & 0.14 & 0.44 & 0.98 & 0.82 & 0.28 & 0.66 & 0.55 \\
& Prompt 4 & 0.00 & 0.30 & 1.00 & 0.86 & 0.22 & 0.56 & 0.49 \\
& Prompt 5 & 0.00 & 0.02 & 0.76 & 0.28 & 0.00 & 0.16 & 0.20 \\
\midrule
\multirow{5}{*}{Qwen 80B} 
 & Prompt 1 & 0.00 & 0.00 & 0.00 & 0.00 & 0.00 & 0.00 & 0.00 \\
 & Prompt 2 & 0.00 & 0.00 & 0.00 & 0.00 & 0.00 & 0.00 & 0.00 \\
 & Prompt 3 & 0.00 & 0.00 & 0.00 & 0.00 & 0.00 & 0.00 & 0.00 \\
 & Prompt 4 & 0.00 & 0.00 & 0.00 & 0.00 & 0.00 & 0.00 & 0.00 \\
 & Prompt 5 & 0.00 & 0.00 & 0.00 & 0.00 & 0.00 & 0.00 & 0.00 \\
\bottomrule
\end{tabular}
\caption{Error rate of each End-to-end LLM for each action for the five prompts we used.}
\label{tab:error_rates}
\end{table*}

\subsection{Extended Generalization Results}
\label{sec: app_res_generalization}

\subsubsection{Generalization Alignment Results}
\label{sec:alignment_results_gen}

Table~\ref{tab:alignment_method1} reports the alignment rates of the \textsc{COMETH} pipeline when applied with different LLMs across six moral actions and three prompts. Overall, the results demonstrate that grounding predictions in structured action representations substantially improves generalization compared to end-to-end LLM approaches. All three models—Llama 8B, Mistral 8B, and Qwen 80B—achieve mean alignment rates around 0.57–0.63 using our method, effectively doubling the performance of direct LLM outputs.

Notably, the \textsc{COMETH} pipeline shows consistent improvements across prompts and actions. Mistral 8B and Qwen 80B reach the highest mean alignment rates (0.63), highlighting the robustness of the approach even when feature extraction varies slightly. Individual actions such as “Kill to Protect” and “Euthanasia” are consistently well-aligned, indicating that \textsc{COMETH} captures semantically meaningful distinctions that correspond to human moral judgments.

These findings confirm that structured representations and probabilistic context learning provide a reliable and reproducible framework for generalizing human moral evaluations, enhancing both the accuracy and interpretability of small and medium-sized LLMs.

\begin{table*}[t]
\centering
\begin{tabular}{llccccccc}
\toprule
\textbf{LLM} & \textbf{Prompt} & \textbf{Lie by Interest} & \textbf{Lie to Support} & \textbf{Euthanasia} & \textbf{Kill to Protect} & \textbf{Steal} & \textbf{Protest} & \textbf{Mean} \\
\midrule
\multirow{3}{*}{Llama 8B} 
& Prompt 1 & 0.30 & 0.24 & 0.78 & 0.84 & 0.84 & 0.50 & 0.58 \\
& Prompt 2 & 0.34 & 0.30 & 0.76 & 0.88 & 0.88 & 0.38 & 0.59 \\
& Prompt 3 & 0.22 & 0.40 & 0.66 & 0.82 & 0.84 & 0.34 & 0.55 \\
\midrule
\multirow{3}{*}{Mistral 8B} 
& Prompt 1 & 0.26 & 0.50 & 0.66 & 0.90 & 0.82 & 0.42 & 0.59 \\
& Prompt 2 & 0.36 & 0.36 & 0.70 & 0.94 & 0.90 & 0.54 & 0.63 \\
& Prompt 3 & 0.38 & 0.20 & 0.66 & 0.94 & 0.82 & 0.46 & 0.58 \\
\midrule
\multirow{3}{*}{Qwen 80B} 
& Prompt 1 & 0.32 & 0.30 & 0.64 & 0.90 & 0.76 & 0.56 & 0.58 \\
& Prompt 2 & 0.36 & 0.40 & 0.76 & 0.90 & 0.82 & 0.54 & 0.63 \\
& Prompt 3 & 0.26 & 0.34 & 0.62 & 0.92 & 0.78 & 0.52 & 0.57 \\
\bottomrule
\end{tabular}
\caption{\textbf{Alignment rates of the \textsc{COMETH} pipeline.} Results are shown per prompt for each LLM across actions. Compared to the end-to-end LLM baselines (which average $\sim$30\% alignment), \textsc{COMETH} effectively doubles the predictive alignment with human moral judgments across all tested models.}
\label{tab:alignment_method1}
\end{table*}

\subsubsection{Example of features evaluation for scenarios belonging to the same cluster.}

Table \ref{tab:compact_examples} shows an example of features evaluation by \textsc{COMETH}'s generalization module for scenarios belonging to the same cluster. It shows that the evaluations of the features by the LLMs we used is coherent with the one a human would do.

\begin{table}[t] 
\centering
\begin{tabular}{lcc} 
\toprule
\textbf{Feature} & \textbf{Scen. A} & \textbf{Scen. B} \\
\midrule
Exposing corruption via data  & 0 & 0 \\
Looting during protest        & 0 & 0 \\
Military supply theft         & 0 & 1 \\
Bicycle for transport         & 0 & 0 \\
Communications gear           & 0 & 0 \\
During civil unrest           & 0 & 1 \\
During disaster               & 1 & 1 \\
Firewood from neighbor        & 0 & 0 \\
Food for family               & 1 & 0 \\
Personal gain                 & 0 & 0 \\
Urgent need                   & 1 & 1 \\
From corporate funds          & 0 & 0 \\
From neighbors                & 1 & 0 \\
From warehouses               & 1 & 1 \\
Fruit from stall              & 1 & 0 \\
Laptops to sell               & 0 & 0 \\
Login credentials             & 0 & 0 \\
Medicine for child            & 0 & 0 \\
School supplies               & 0 & 0 \\
Smartphone after hours        & 0 & 0 \\
Snacks from cafeteria         & 1 & 0 \\
Tools for creation            & 0 & 0 \\
Toys from peers               & 0 & 0 \\
Water from enemies            & 0 & 0 \\
Winter essentials             & 1 & 1 \\
\bottomrule
\end{tabular}
\caption{Example of binary feature extraction by the \textsc{COMETH} generalization module. \textbf{Scen. A}: A father steals food to feed his starving children. \textbf{Scen. B}: A soldier takes rations from enemy camps to feed his platoon.}
\label{tab:compact_examples}
\end{table}

\subsubsection{Features weights Example}
\label{sec:features_examples}

\begin{figure}[!t]
    \centering
    \includegraphics[width=0.95\columnwidth]{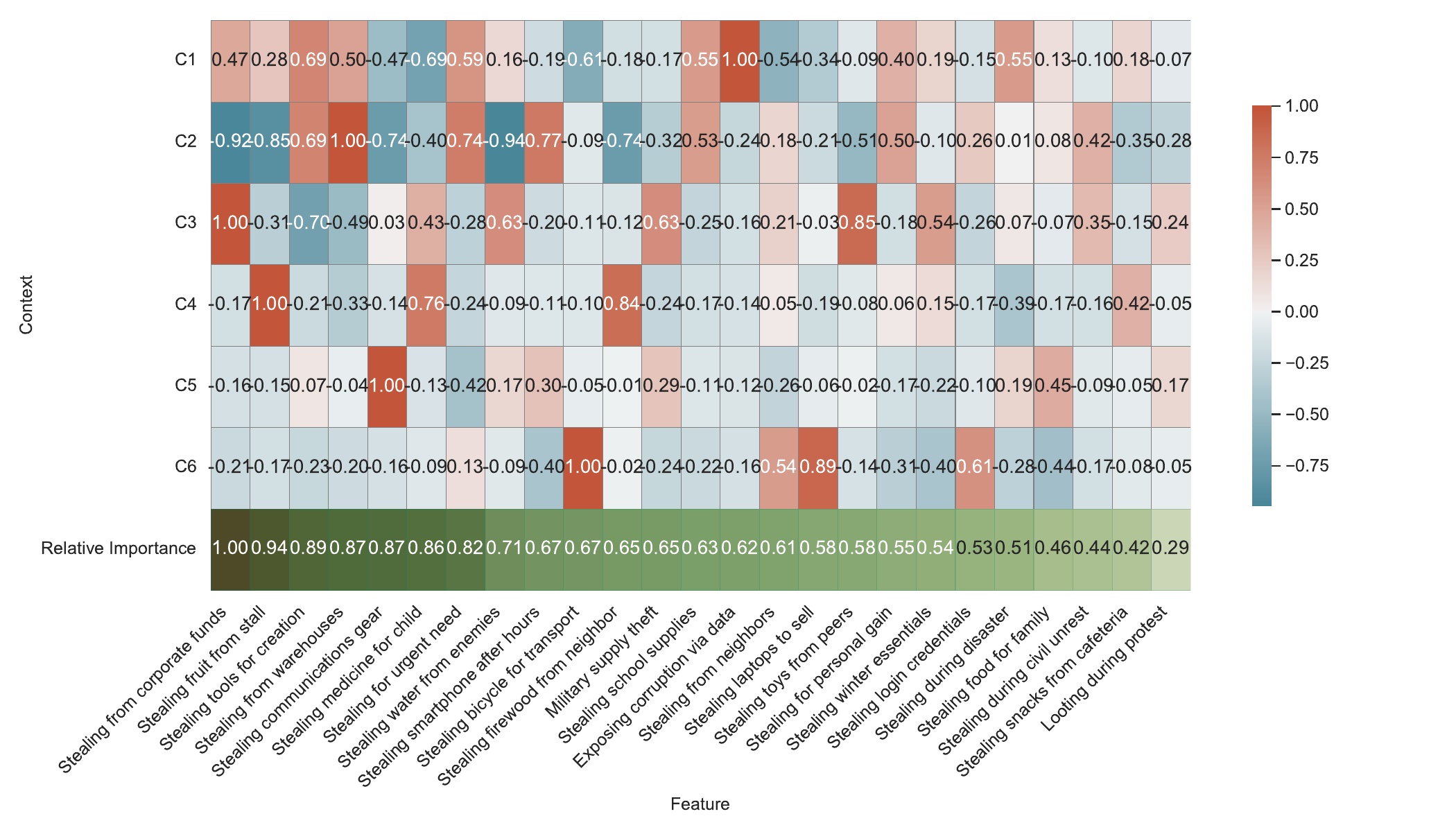}
    \caption{\textbf{\textsc{COMETH} feature weights obtained for the action "Steal"} (powered by Qwen-3-Next-80B-A3B-Instruct). 
    The plot reports the relative importance of each feature in shaping cluster assignments, highlighting how different attributes influence whether the scenario is more likely to be assigned to each cluster and then judged with support or blame.
}
    \label{fig: features2}
\end{figure}

\begin{figure}[!t]
    \centering
    \includegraphics[width=0.95\columnwidth]{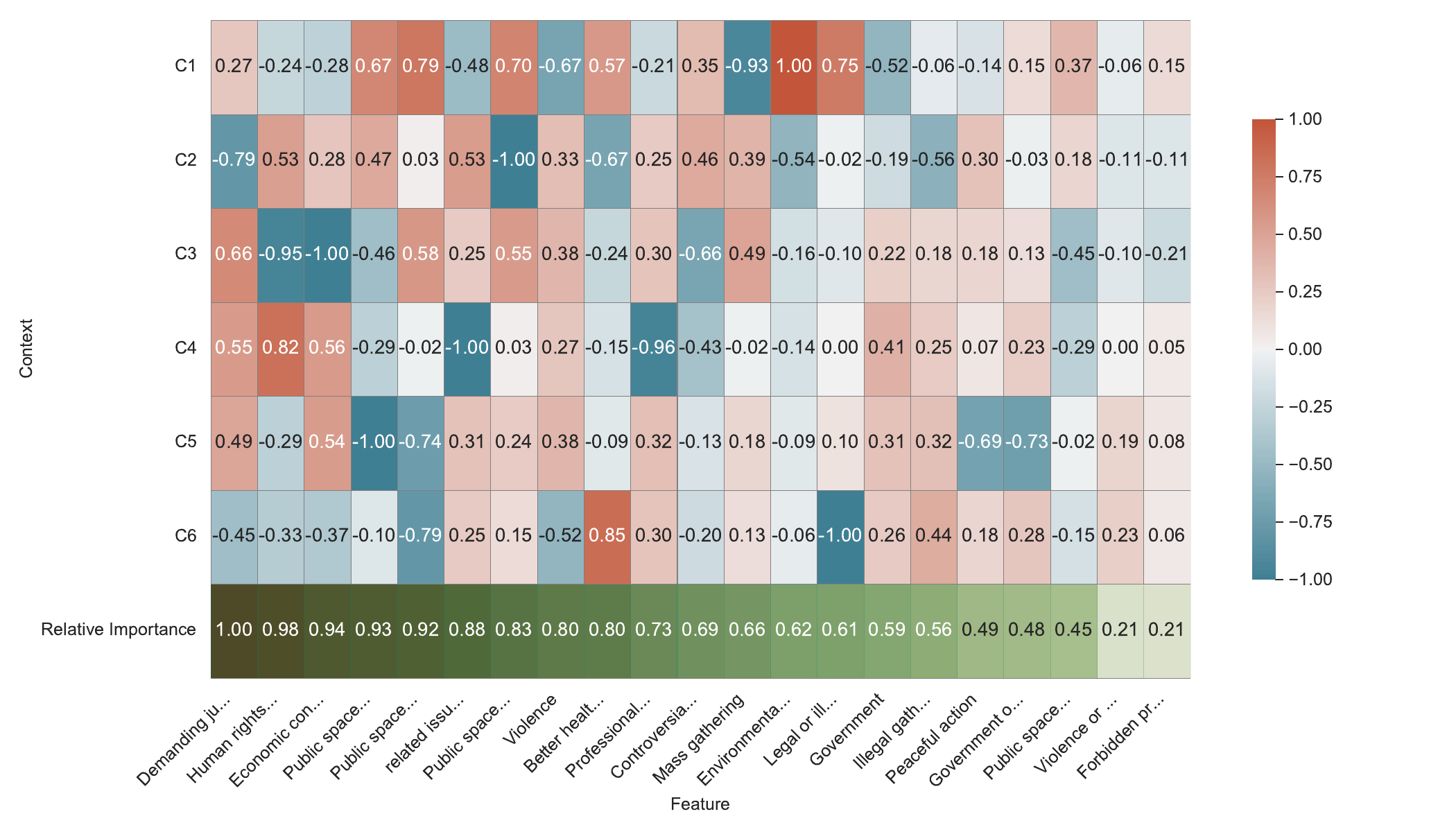}
    \caption{\textbf{\textsc{COMETH} feature weights obtained for the action "Participate in an Illegal Protest"} (powered by Mistral8B-Instruct). 
    The plot reports the relative importance of each feature in shaping cluster assignments, highlighting how different attributes influence whether the scenario is more likely to be assigned to each cluster and then judged with support or blame.
}
    \label{fig: features3}
\end{figure}

\begin{figure}[!t]
    \centering
    \includegraphics[width=0.95\columnwidth]{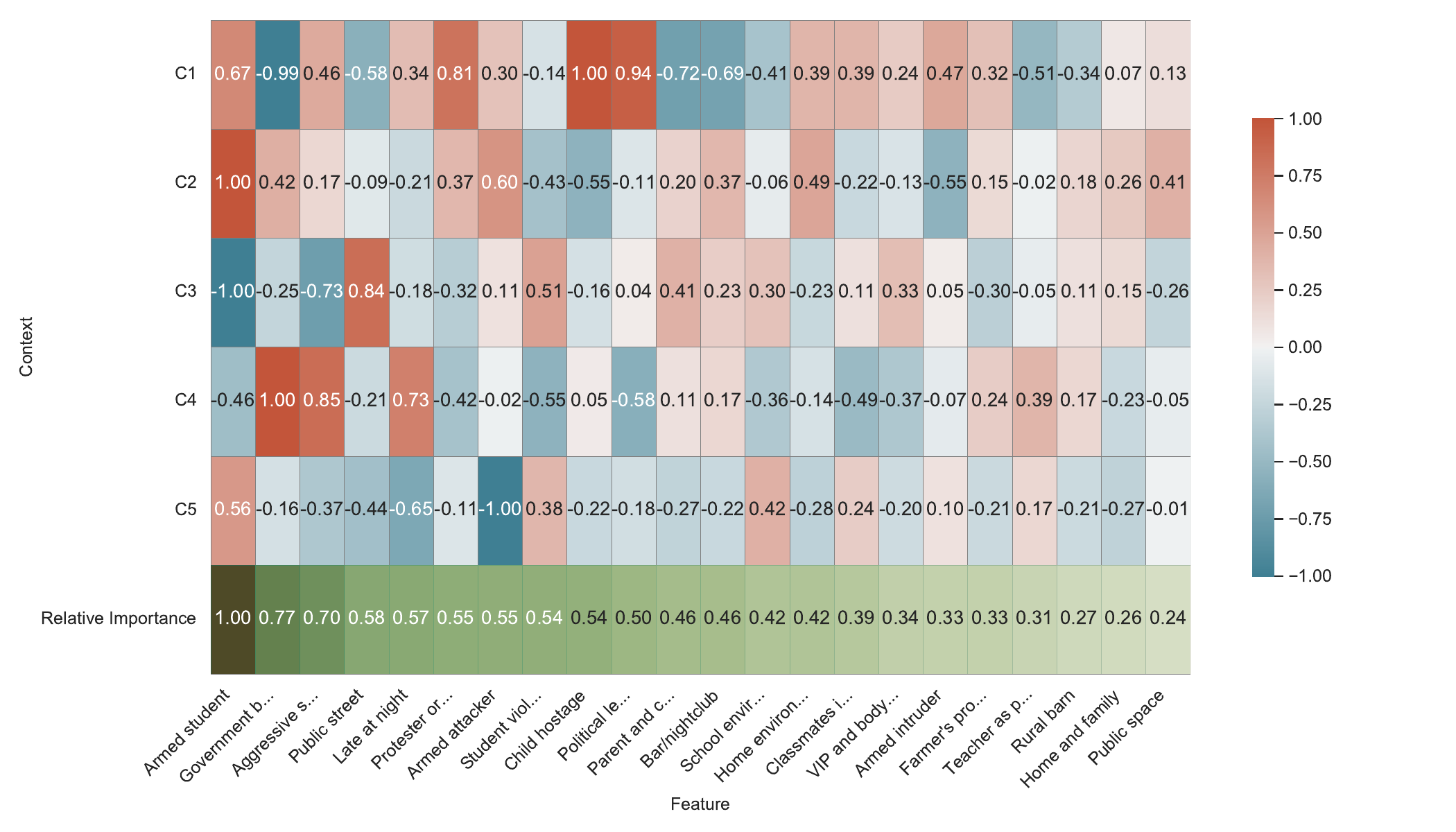}
    \caption{\textbf{\textsc{COMETH} feature weights obtained for the action "Kill to Protect"} (powered by Llama8B-Instruct). 
    The plot reports the relative importance of each feature in shaping cluster assignments, highlighting how different attributes influence whether the scenario is more likely to be assigned to each cluster and then judged with support or blame.
}
    \label{fig: features4}
\end{figure}

In this subsection we show other examples of Feature weights inferred by \textsc{COMETH} for different prompts, actions, and models. These weights are shown in Figures \ref{fig: features2}, \ref{fig: features3}, and \ref{fig: features4}. 

\section{Ablation tests}

\subsection{Pre-processing Ablation Studies}

We conducted a series of ablation experiments to assess the robustness of the COMETH pre-processing pipeline and to evaluate the necessity of its main design choices. In particular, we examined (i) the impact of fixing the number of clusters in K-means and (ii) the role of LLM-based scenario reformulation prior to clustering.

\subsubsection{Silhouette-based Selection of the Number of Clusters}

In our main experiments, the number of clusters in K-means was fixed to six, corresponding to the six ideal core actions underlying the scenario set. As an ablation, we instead selected the number of clusters automatically using the silhouette score, without enforcing any prior constraint on $k$.

Across 15 data variants (3 different LLMs and 5 prompts), silhouette-based selection resulted in a variable number of clusters, ranging from 7 to 14. Quantitative alignment with the ideal clustering, measured using ARI, NMI, and V-measure, decreased compared to the fixed-$k$ setting (see Table~\ref{tab:silhouette_metrics}). Representative results are reported below:

\begin{table}[!t] 
\centering
\setlength{\tabcolsep}{4pt} 
\begin{tabular}{llcccc}
\toprule
\textbf{LLM} & \textbf{Prompt} & \textbf{$k_{\text{optimal}}$} & \textbf{ARI} & \textbf{NMI} & \textbf{V-measure} \\
\midrule
Mistral 7B & Minimalist & 13 & 0.550 & 0.680 & 0.680 \\
Mistral 7B & Infinitive & 13 & 0.546 & 0.695 & 0.695 \\
Mistral 7B & MainAct    & 9  & 0.651 & 0.719 & 0.719 \\
Mistral 7B & OneWord    & 14 & 0.523 & 0.690 & 0.690 \\
Mistral 7B & NounPhrase & 7  & 0.818 & 0.836 & 0.836 \\
\midrule
Llama 8B   & Minimalist & 9  & 0.666 & 0.759 & 0.759 \\
Llama 8B   & Infinitive & 14 & 0.540 & 0.694 & 0.694 \\
Llama 8B   & MainAct    & 9  & 0.714 & 0.742 & 0.742 \\
Llama 8B   & OneWord    & 14 & 0.383 & 0.616 & 0.616 \\
Llama 8B   & NounPhrase & 8  & 0.666 & 0.726 & 0.726 \\
\midrule
Qwen 80B   & Minimalist & 14 & 0.631 & 0.780 & 0.780 \\
Qwen 80B   & Infinitive & 14 & 0.616 & 0.752 & 0.752 \\
Qwen 80B   & MainAct    & 12 & 0.716 & 0.804 & 0.804 \\
Qwen 80B   & OneWord    & 14 & 0.546 & 0.714 & 0.714 \\
Qwen 80B   & NounPhrase & 13 & 0.658 & 0.771 & 0.771 \\
\bottomrule
\end{tabular}
\caption{\textbf{Silhouette-based clustering results across LLMs and prompts.} For each LLM and prompt variant, we report the optimal number of clusters selected by the silhouette score ($k_{\text{optimal}}$) and the corresponding clustering alignment with the ideal core-action partition, measured using Adjusted Rand Index (ARI), Normalized Mutual Information (NMI), and V-measure. Although $k_{\text{optimal}}$ varies across configurations, the resulting clusters generally correspond to semantically coherent subdivisions of the ideal core-action clusters.}
\label{tab:silhouette_metrics}
\end{table}

\begin{figure}[!t] 
    \centering
    \includegraphics[width=\columnwidth]{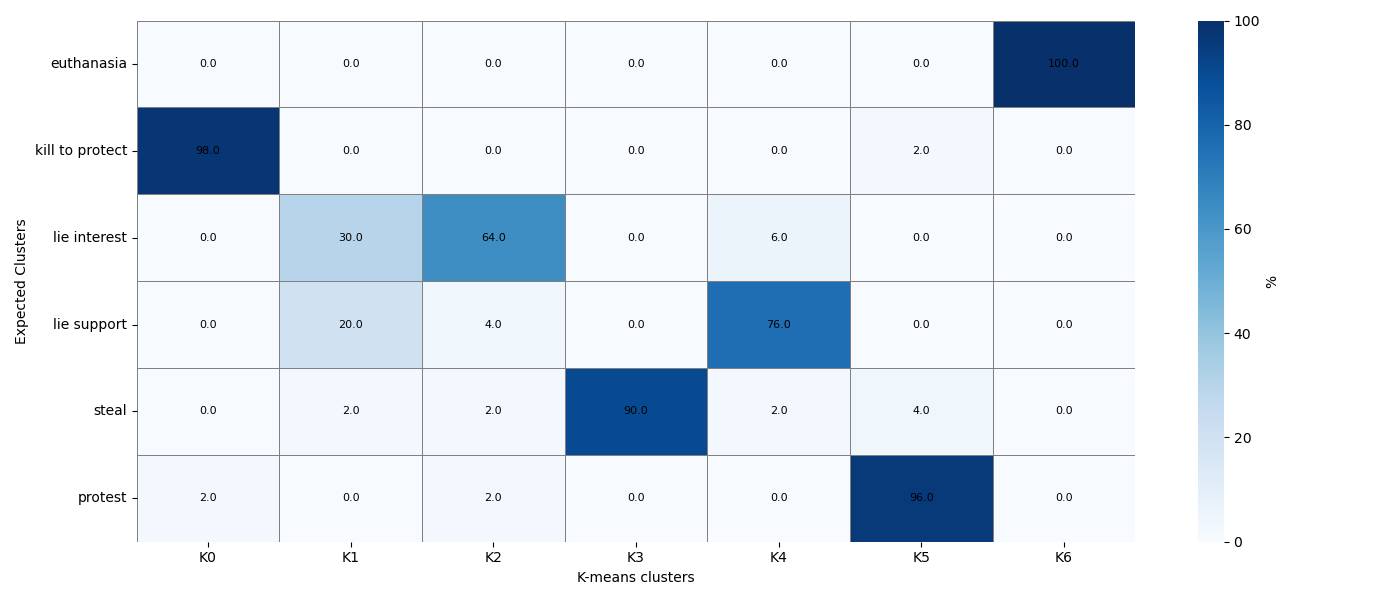}
    \caption{\textbf{Confusion matrix between the ideal clustering and a K-means clustering.} K-means clustering on the embeddings of the reformulation of the scenarios by Mistral 7B with the NounPhrase prompt.}
    \label{fig:conf_matrix}
\end{figure}

Although the resulting number of clusters does not match the expected six, qualitative inspection reveals that the obtained clusters are largely subdivisions of the ideal core-action clusters rather than mixtures of distinct actions. This is illustrated in Figure~\ref{fig:conf_matrix}, where scenarios corresponding to different core actions are rarely grouped together.

These results indicate that fixing the number of clusters is beneficial in our current proof-of-concept setting, where the target structure is known. Nevertheless, the silhouette-based approach remains viable and produces semantically coherent clusters, suggesting that COMETH can generalize to settings where the number of underlying actions is unknown or where a larger and more diverse scenario set would naturally support finer-grained clustering.

\subsubsection{Clustering Without LLM-based Reformulation}

We further evaluated the necessity of LLM-based pre-processing by performing K-means clustering directly on the original scenario embeddings, without any LLM reformulation.

\begin{figure}[!t] 
    \centering
    \includegraphics[width=\columnwidth]{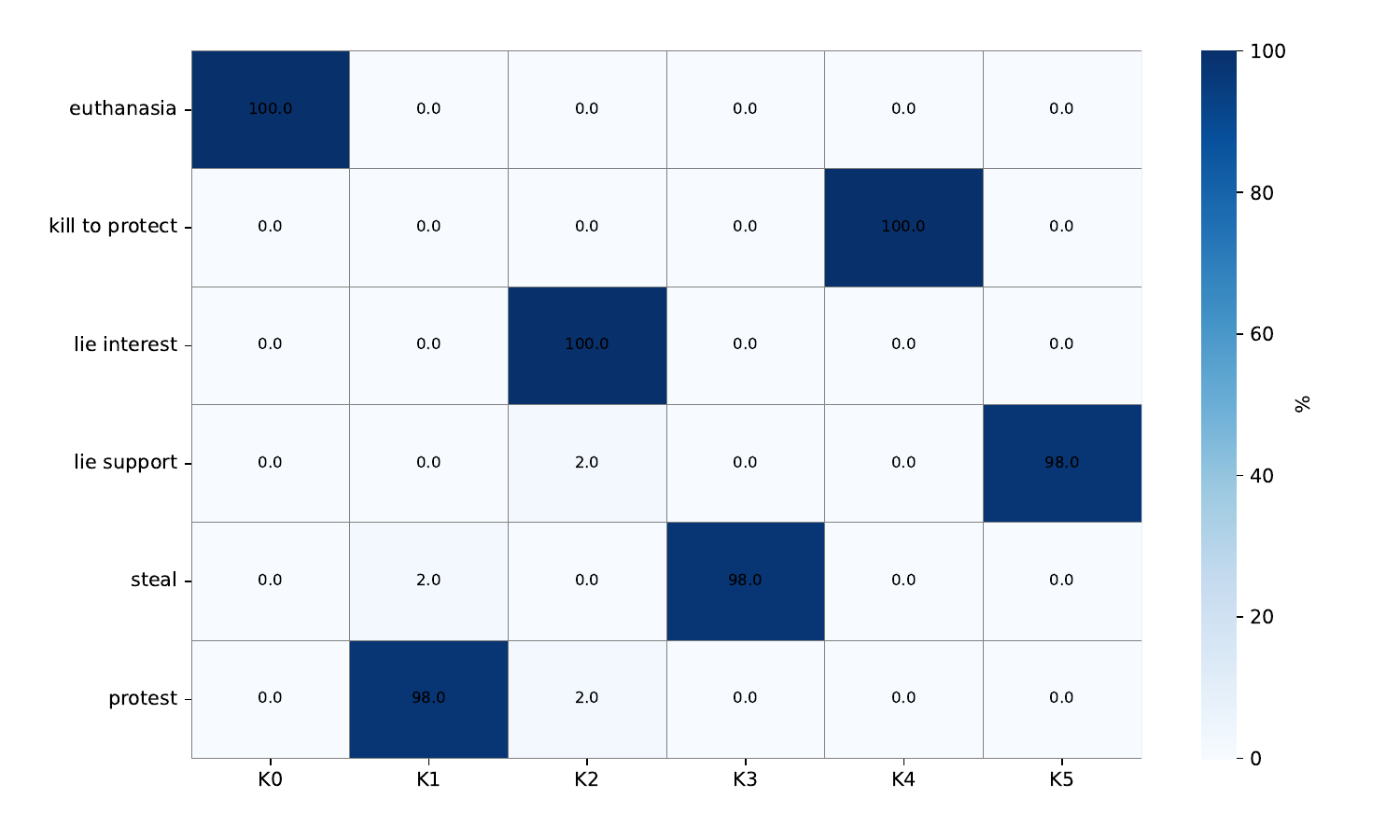}
    \caption{\textbf{K-means clustering on the scenarios embeddings with a fix number of clusters}}
    \label{fig:without_llms_6}
    
    \vspace{0.5cm} 
    
    \includegraphics[width=\columnwidth]{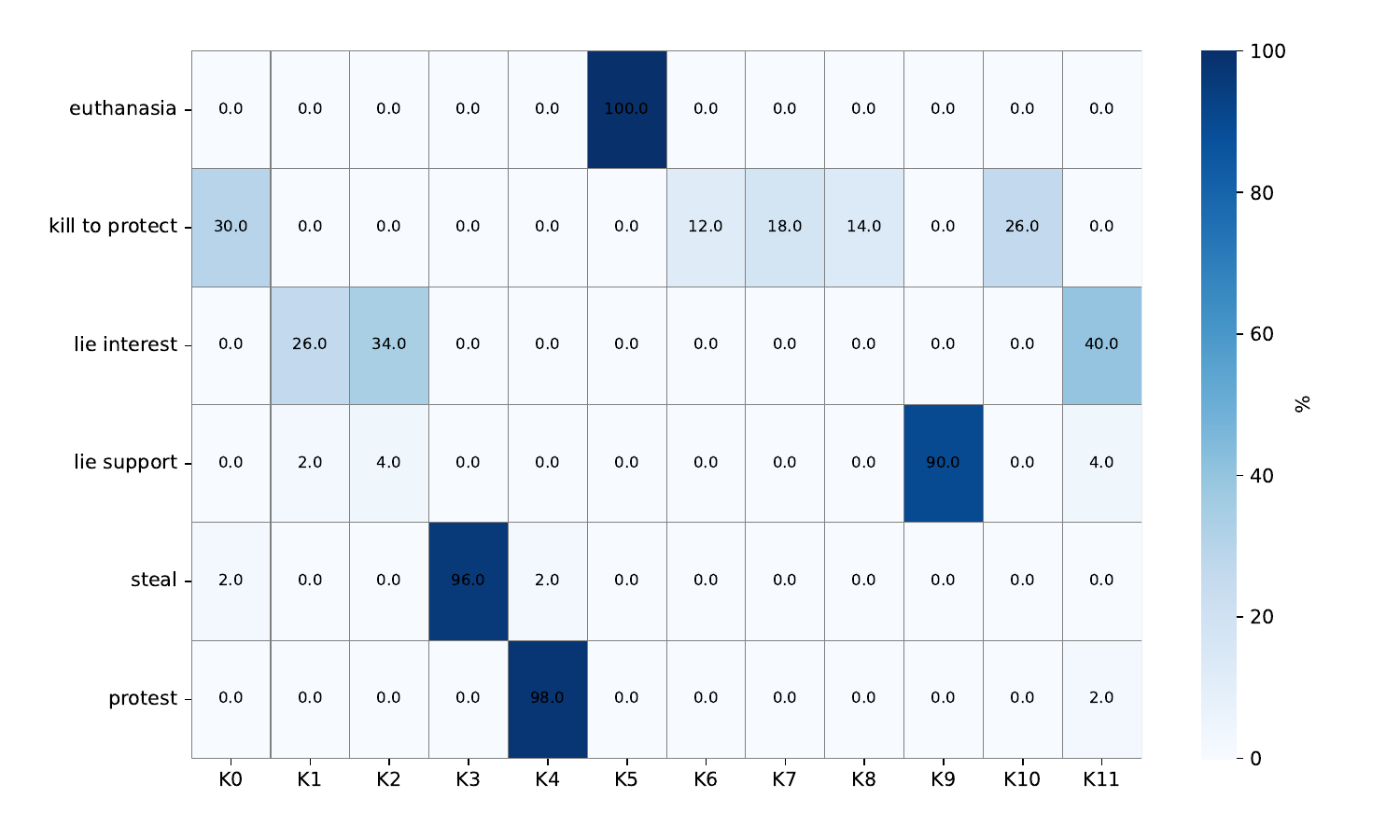}
    \caption{\textbf{K-means clustering on the scenarios embeddings with a number of clusters based on the silhouette score}}
    \label{fig:without_llms}
\end{figure}

When the number of clusters was fixed to six, clustering performance was nearly perfect, achieving ARI = 0.976, NMI = 0.973, and V-measure = 0.973 (Figure~\ref{fig:without_llms_6}). When the number of clusters was selected using the silhouette score (yielding $k=12$), performance decreased (ARI = 0.786, NMI = 0.837, V-measure = 0.837), but clusters remained homogeneous and interpretable as subdivisions of single core actions (Figure~\ref{fig:without_llms}).

These results show that, for the present dataset, LLM-based reformulation is not strictly required to recover the underlying core actions. However, this finding should be interpreted with caution: the scenarios in our dataset follow relatively uniform linguistic and structural patterns. In more diverse or complex scenario collections, surface-level embeddings may no longer suffice, and LLM-based abstraction may become critical to align scenarios along action-level semantics.

\subsubsection{Summary}

Overall, these ablation studies demonstrate that COMETH’s pre-processing pipeline is robust to variations in both cluster-number selection and the use of LLM-based reformulation. While fixing $k$ and using LLMs yields the most stable results in our current setting, the method remains effective without these constraints, supporting its applicability beyond controlled proof-of-concept scenarios.

\subsection{Clustering K-means vs Probabilistic Context Learner}

\begin{figure}[!t]
    \centering
    \includegraphics[width=\columnwidth]{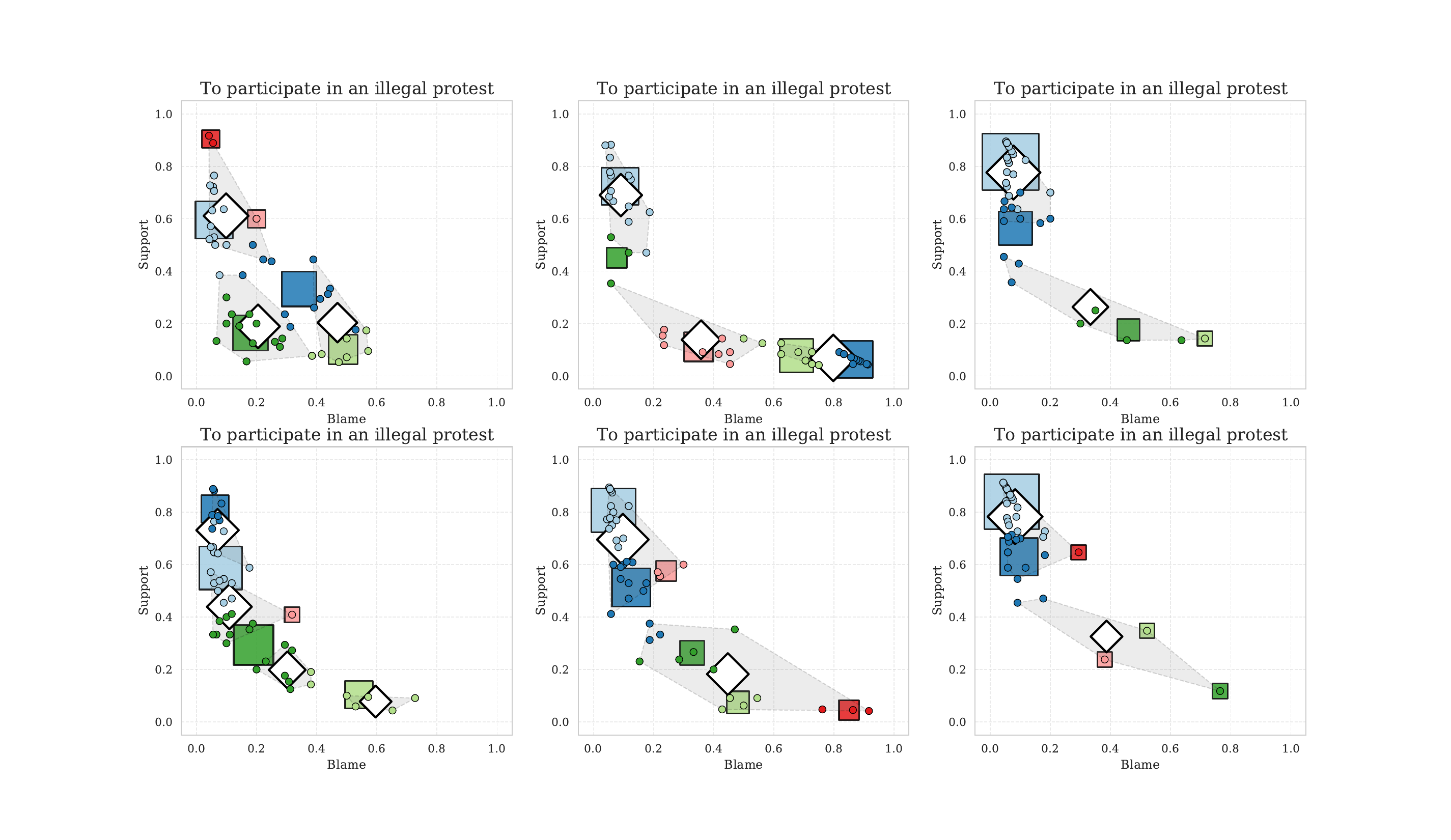}
    \caption{\textbf{Comparison of the two clusterings.} Dots represent the scenarios, squares represent the Probabilistic Context Learner clusters, and diamonds represent K-means clusters. The color of the dots depends on the Probabilistic Context Learner cluster they belong to. Grey areas surround the scenarios belonging to each K-means clusters. }
    \label{fig:two_clusterings}
\end{figure}

We performed a static K-means clustering to serve as a baseline for evaluating our Probabilistic Context Learner. For K-means, the number of clusters per action was determined using the silhouette score.

As illustrated in Figure \ref{fig:two_clusterings}, the resulting cluster partitions differ substantially between the two approaches. K-means produces fewer clusters overall, leading to a coarser representation of contextual variations. This reduced granularity causes the loss of meaningful distinctions — for example, scenarios considered immoral by 40\% versus 80\% of participants are often grouped together, despite representing clearly different judgment tendencies.

\begin{figure}[!t] 
    \centering
    \includegraphics[width=0.7\columnwidth]{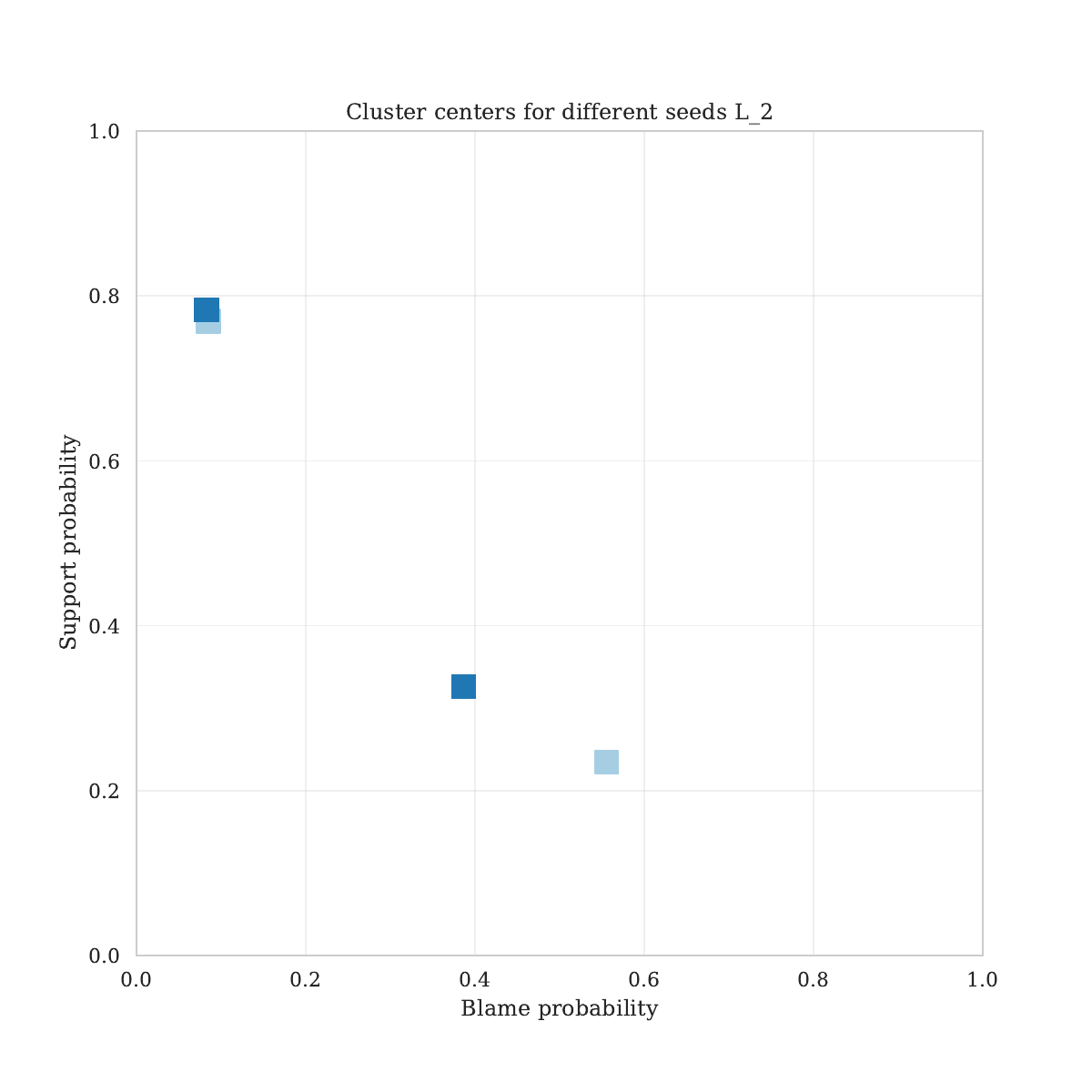}
    \caption{\textbf{Stability of the K-means clustering.} Two examples of clusterings for the same action with the same scenarios but with different seeds showing the lack of stability of the k-means clustering.}
    \label{fig:robustness_k_means}
\end{figure}

Moreover, K-means exhibits instability across different random initializations: the cluster assignments vary significantly depending on the seed, as shown in Figure \ref{fig:robustness_k_means}. In contrast, COMETH yields a substantially more stable clustering (see the Extended Pre-processing Results section), reinforcing its robustness as a modeling tool.

Finally, both clustering methods achieve similar alignment rates — defined as the proportion of scenarios in a cluster whose dominant behavioral response matches the cluster’s overall distribution — as reported in Table \ref{tab:table_comparison_k_means_cometh}. Therefore, the main advantage of COMETH lies not in alignment quality, but in its improved granularity and stability, enabling more precise interpretation of moral judgment variability across contexts. The Probabilistic Context Learner uncovers stable, fine-grained and psychologically meaningful structure in human moral judgments that K-means fails to capture, demonstrating the necessity of probabilistic context modeling beyond distance-based clustering.

\begin{table}[!t] 
\centering
\setlength{\tabcolsep}{3.5pt} 
\begin{tabular}{l c c c c}
\toprule
 & \multicolumn{2}{c}{\textbf{K-Means}} & \multicolumn{2}{c}{\textbf{COMETH}} \\ 
\cmidrule(lr){2-3} \cmidrule(lr){4-5}
\textbf{Action} & \textbf{Align. (\%)} & \textbf{Clusters} & \textbf{Align. (\%)} & \textbf{Clusters} \\
\midrule
Practice Euthanasia & 94 & 2 & 90 & 4 \\
Kill to Protect     & 78 & 4 & 94 & 5 \\
Lie by Interest     & 92 & 3 & 92 & 5 \\
Lie to Support      & 96 & 3 & 78 & 6 \\
Steal               & 84 & 2 & 86 & 6 \\
Illegal Protest     & 96 & 2 & 98 & 6 \\
\bottomrule
\end{tabular}
\caption{Comparison of clustering results between K-Means and COMETH across different actions.}
\label{tab:table_comparison_k_means_cometh}
\end{table}

\subsection{Generalization ablations}

\subsubsection{Generalization without learning phase}

We evaluated COMETH using only the feature frequencies, without performing the weight-learning phase, while keeping the same k-fold cross-validation procedure. Across all actions, prompts, and LLMs, the resulting alignment rates were substantially lower, typically 2 to 3 times lower than those obtained with the full learning phase (see Table \ref{tab:method2_aligned_mean}).

\begin{table}[!t] 
\centering
\small
\setlength{\tabcolsep}{3.5pt}
\begin{tabular}{l c c c c c c c}
\toprule
\textbf{Model} & \textbf{D\_1} & \textbf{D\_2} & \textbf{K\_1} & \textbf{K\_2} & \textbf{L\_1} & \textbf{L\_2} & \textbf{Mean} \\
\midrule
\multicolumn{8}{c}{\textit{Prompt 1}} \\
\midrule
Llama 8B   & 0.22 & 0.16 & 0.46 & 0.20 & 0.40 & 0.10 & 0.2567 \\
Mistral 7B & 0.28 & 0.34 & 0.28 & 0.28 & 0.12 & 0.20 & 0.2500 \\
Qwen 80B   & 0.06 & 0.32 & 0.30 & 0.36 & 0.18 & 0.38 & 0.2667 \\
\midrule
\multicolumn{8}{c}{\textit{Prompt 2}} \\
\midrule
Llama 8B   & 0.02 & 0.40 & 0.28 & 0.12 & 0.26 & 0.26 & 0.2233 \\
Mistral 7B & 0.32 & 0.16 & 0.34 & 0.14 & 0.26 & 0.54 & 0.2867 \\
Qwen 80B   & 0.14 & 0.52 & 0.46 & 0.42 & 0.44 & 0.42 & 0.4000 \\
\midrule
\multicolumn{8}{c}{\textit{Prompt 3}} \\
\midrule
Llama 8B   & 0.12 & 0.08 & 0.42 & 0.22 & 0.10 & 0.10 & 0.1700 \\
Mistral 7B & 0.20 & 0.14 & 0.36 & 0.26 & 0.16 & 0.42 & 0.2600 \\
Qwen 80B   & 0.18 & 0.22 & 0.38 & 0.28 & 0.38 & 0.40 & 0.3100 \\
\bottomrule
\end{tabular}
\caption{Aligned scores of the generalization across prompts and models for different datasets, without learning.}
\label{tab:method2_aligned_mean}
\end{table}

This experiment demonstrates that the weight-learning step is crucial for COMETH’s performance, highlighting the effectiveness and necessity of our method in capturing the relative importance of contextual features for accurate moral alignment.

\subsubsection{Generalization with building features}

We also evaluated the generalization using the hand-crafted buildicontextual features which correspond to contextual elements we assigned by hand when creating the scenarios. The results show high alignment rates for some actions (namely, Practice Euthanasia or Participate in an Illegal Protest), while other actions achieved lower performance (see Table \ref{tab:action_scores}). This indicates that LLM-extracted features can be highly informative for certain actions, but may be less informative than the hand-crafted features for others.

\begin{table}[!t] 
\centering
\begin{tabular}{lcc}
\toprule
\textbf{Action} & \textbf{Accuracy} & \textbf{Aligned} \\
\midrule
Practice Euthanasia & 0.62 & 0.90 \\
Kill to Protect & 0.36 & 0.44 \\
Lie by Interest & 0.08 & 0.20 \\
Lie to Support & 0.14 & 0.32 \\
Steal & 0.38 & 0.74 \\
Illegal Protest & 0.68 & 0.94 \\
\bottomrule
\end{tabular}
\caption{Accuracy and Aligned scores for each action obtained with the COMETH pipeline by using hand-crafted contextual features during the Generalization phase.}
\label{tab:action_scores}
\end{table}

Although hand-crafted features are not available in real-world applications, this experiment demonstrates that high alignment rates are achievable given high-quality features. These findings motivate future work on improving the consistency and quality of LLM-based feature extraction to maximize COMETH’s generalization across all action types.

\bibliography{aaai2026}

\end{document}